\documentclass[10pt, conference, letterpaper]{IEEEtran}

\IEEEoverridecommandlockouts                              % This command is only needed if 
\usepackage{cite}
\usepackage{subfigure}
\usepackage{url}
\usepackage{verbatim,epsf}
\usepackage[font=footnotesize]{subfig}

\usepackage{multicol}
\usepackage{multirow}
\usepackage[font=footnotesize,labelfont=bf]{
	caption}

\usepackage{siunitx}
\usepackage{bm}
\usepackage{tabularx}
\usepackage{balance} %MAU this is needed to balance the column of the last page

\usepackage{amssymb}
\usepackage{amsmath}
\usepackage{graphicx}
\usepackage{epstopdf}

\usepackage{epsfig, color}

\usepackage{comment} % Package per i commenti
\usepackage{amsfonts} % Package per i caratteri  matematici spciali
\usepackage{amsmath} % Package per i caratteri  matematici spciali
\usepackage{rotating}

\usepackage{amssymb}
\usepackage{amsmath}
\usepackage{amsfonts}
\usepackage{wasysym}

\def\QED{\mbox{\rule[0pt]{1.5ex}{1.5ex}}}

\usepackage{algorithm}
\usepackage{algorithmic}

\usepackage{color}

\begin{document}
\title{%\huge
	{3D Reconstruction of Multiple Objects by \\
		mmWave Radar on UAV}}
%Incentive Mechanism for Edge Computing via Shapley-value} 
\author{ 
\IEEEauthorblockN{Yue Sun\IEEEauthorrefmark{1},
	Zhuoming Huang\IEEEauthorrefmark{2},  Honggang Zhang\IEEEauthorrefmark{2}, Xiaohui Liang\IEEEauthorrefmark{1}\\
	\IEEEauthorrefmark{1} Computer Science Dept., UMass Boston, Boston, MA. \\
		\IEEEauthorrefmark{2} Engineering Dept., UMass Boston, Boston, MA. \\
	Email: \{yue.sun001, zhuoming.huang001, honggang.zhang, xiaohui.liang\}@umb.edu
	%\IEEEauthorrefmark{2} UMass Lowell, Lowell, MA. Email: bliu@cs.uml.edu
}
}

\maketitle
\thispagestyle{empty}
\pagestyle{empty}

%%%%%%%%%%%%%%%%%%%%%%%%%%%%%%%%%%%%%%%%%%%%%%%%%%%%%%%%%%%%%%%%%%%%%%%%%%%%%%%%
\begin{abstract}
In this paper, we explore the feasibility of utilizing a mmWave radar sensor installed on a UAV
to reconstruct the 3D shapes of multiple objects in a space.
The UAV hovers at various locations in the space, and its onboard radar senor 
collects raw radar data via scanning the space with Synthetic Aperture Radar (SAR) operation.
%due to the low resolution of commodity radar sensors. 
%in an environment, and then based on the signals to 
The radar data is sent to a deep neural network model, which 
%We design a three-stage deep neural network architecture to take as input the raw radar signals 
%and 
outputs the point cloud reconstruction of the multiple objects in the space. 
% is needed as the commodity radar sensors are of low resolution. However, 
%due to the instability of 
We evaluate two different models. Model 1 is  
our recently proposed 3DRIMR/R2P model \cite{sun2022icann}, and Model 2 is formed by adding a segmentation stage 
in the processing pipeline of Model 1.
%The first stage of our network architecture generates 2D depth images of the environment based radar
%signals, the second stage conducts scene segmentation to get the depth images of individual objects, 
%and the last stage generates 3D point clouds of those objects based on their depth images. 
Our experiments have demonstrated that both models are promising in solving the multiple object reconstruction problem. 
We also show that Model 2, despite producing denser and smoother point clouds,
%sometimes producing visually more pleasing results, 
can lead to higher reconstruction loss or even loss of objects. 
In addition, we find that both models are robust to the highly noisy radar data obtained by unstable SAR operation 
due to the instability or vibration of a small UAV hovering at its intended scanning point. 
Our exploratory study has shown a promising direction of applying mmWave radar sensing in 3D object reconstruction. 
% and possibly 
%many incorrect and/or inconsistent points generated during the workflow of the architecture. 
%Since a small UAV is usually not perfectly stable while hovering, the SAR operation of the onboard radar sensors 

\end{abstract}

% ================================================================================================================
\section{INTRODUCTION}

Frequency Modulated Continuous Wave (FMCW) Millimeter Wave (mmWave) radar sensing 
recently has been shown as an effective sensing tool in low visibility 
environment, thus making it a promising sensing technique in autonomous vehicles \cite{HawkEye}
and search/rescue scenarios \cite{mobisys20smoke}. 
The capability of 3D object reconstruction is important in a search and rescue scenario, e.g., firefighting scenes, 
where heavy smoke makes optical sensing not practical.  
However it is quite challenging to reconstruct 3D object shapes based on
mmWave radar data because the data is usually of low resolution, sparsity, specularity, 
and large noise due to multi-path effects. 
Recent work on 3D reconstruction has made some progress 
in this direction \cite{HawkEye, mobisys20smoke, sun20213drimr, sun2022icann}, 
focusing on single object reconstruction. 
%Our work in \cite{sun20213drimr} introduce a 2-stage architecture to  

In this paper, we go one step further to explore the feasibility of 
reconstructing 3D shapes of multiple objects in a space (which is more challenging than single object reconstruction), based on 
mmWave radar data collected from a sensor mounted on a UAV. 
We let the UAV fly in the space and hover at various locations to scan/collect radar signals in the space. 
Then the UAV can obtain a collection of heatmaps or energy intensity maps of the space after FFT processing of the 
received raw radar signals. We take the heatmaps as input to a 
deep neural network model to generate the smooth and dense point clouds
of the multiple objects in the space. We investigate two different deep neural network models for point cloud generation. 
(1) Model 1 is our recently proposed 3DRIMR/R2P model \cite{sun2022icann}.
%to reconstruction of multiple objects. 
It consists of two stages. In stage 1 it generates the 2D depth images of the space based radar
signals. 
In stage 2 it generates 3D point clouds of those objects based on their depth images 
obtained from stage 1.
Model 1 is used to reconstruct single objects in our previous work \cite{sun2022icann}.  
(2) Model 2 is formed by adding an image segmentation stage between stage 1 and stage 2 of Model 1.
The segmentation stage separates the objects in the depth image representation of a scene (result of stage 1), 
so that they can be reconstructed separately in the next stage. The reason of introducing a segmentation stage
is that working on single objects separately is easier than on multiple objects together when generating final 
point clouds. 
%The second stage conducts scene segmentation to get the depth images of individual objects. 

In addition, the sensing/reconstruction system we consider only utilizes 
low-cost commodity mmWave radar sensors (e.g., \cite{iwr6843}) which have low resolution. 
Therefore in order to obtain high resolution radar scan of a space, we mount a light-weight slider mechanism with 
radar sensor on a UAV so that the UAV can conduct  
Synthetic Aperture Radar (SAR) operation by sliding the sensor horizontally and vertically while hovering in the air. 
Due to the hovering instability of a small UAV, the data collected is highly noisy and the heatmaps of 
FFT processing is quite different from those of a stable precise SAR operation. 
Thus we are interested in whether such a vibrating UAV SAR operation can result in any meaningful reconstruction 
in practice.  

Our major contributions are as follows. 
We demonstrate that it is promising to utilize a multi-stage deep neural network model to reconstruct
multiple objects in a space based on mmWave radar data.
In addition, we find that Model 1 has better performance than Model 2 even though 
Model 2 has an extra segmentation stage and can result in denser/smoother point clouds. 
This is because any segmentation error can cause cascading 
failures in the following reconstruction stage, hence making the model less robust and more error-prone. 
%a three-stage deep neural network model that takes as input raw radar data 
%collected from a space where multiple objects are present, and the model generates a 3D reconstruction of the space 
%where those objects are reconstructed in the form of point clouds. 
Furthermore, we find that both models are fairly robust to the 
highly distorted and noisy radar data collected by unstable SAR operation 
due to the vibration/instability of a commodity UAV when hovering.
%However our model is robust to such imperfect SAR data, 
This finding shows that the inherent intricate characteristics of radar energy signature of a space is still retained 
in the imperfect SAR data, and shows that it is feasible to use low-cost small UAV in an
environment sensing/reconstruction mission under low visibility. 
%Our work gives some promising possible direction in applying commodity mmWave radar sensor and UAV to 
%conduct 3D reconstruction in a low visibility environment. 

In the rest of the paper, we briefly discuss related work in Section \ref{sec_related}.
Then we discuss the two models in Section \ref{sec_design}, and 
our experiment results in Section \ref{sec_imp}. 
Finally the paper concludes in Section \ref{sec_conclusion}.

% ================================================================================================================
\section{RELATED WORK}\label{sec_related}

Recently the application of mmWave radar sensing and imaging has been investigated in various areas \cite{mobisys20smoke,HawkEye,vandersmissen2018indoor,yang2020mu,superrf,xue2021mmmesh, sun20213drimr}. 
%particularly in low visibility environment, e.g., \cite{}.
Different from person identification \cite{vandersmissen2018indoor, yang2020mu} and 3D human mesh estimation \cite{xue2021mmmesh}, 
our work aims to reconstruct detailed 3D shapes of various objects in a space. 
Instead of focusing on a specific object (e.g., human body \cite{yang2020mu, xue2021mmmesh}) 
%and relying on some of its inherent characteristics, 
we would like to develop a generic system that can reconstruct the 3D shape of any object. In addition, 
we choose to work on the raw radar energy heatmaps
of a space instead of the point clouds generated by commodity radar sensors as we find them highly sparse and missing lots of information. 

Our work is also closely related to a large body of literature in 3D reconstruction in computer vision \cite{yang20173d,dai2017shape,sharma2016vconv,qi2016pointnet,qi2017pointnet++,yuan2018pcn}.
In particular the design of neural networks used in our architecture is inspired by PointNet \cite{qi2016pointnet}
and PCN \cite{yuan2018pcn}. Our work in this paper is developed based on our recent work on 3D reconstruction of 
a single object from mmWave radar signals \cite{sun20213drimr,sun2022icann}.
%, i.e., the first model is  by introducing a 3-stage design 
%that can segment a radar scanned scene and reconstruct individual objects' shapes in the scene. 
In addition, this work also deals with the input radar energy heatmaps 
%are obtained from Fast Fourier Transformation (FFT) processing 
of highly noisy SAR data due to the hovering instability of a small quadcopter UAV.
%, which is not considered before. 

% ================================================================================================================
%\section{Background}\label{sec_background}
%\subsection{FMCW Millimeter Wave Radar Sensing and Imaging}
%\subsection{Representation of 3D Objects}
%\subsection{Review of 3DRIMR Architecture}

% ================================================================================================================
\section{System Design}\label{sec_design}

\subsection{Overview}
We investigate two different models to address the problem of 3D reconstruction of multiple objects based on mmWave 
radar data. Model 1 is from our recent work 3DRIMR/R2P \cite{sun2022icann}, and 
Model 2 is formed by adding a segmentation stage into Model 1. 
%which can reconstruct the 3D point cloud of a scene, 
%which consists of multiple objects, based on mmWave radar signals captured by UAV.
%The differences between this system and R2P are: 
%(1) mmWave radar signals used in this system have higher resolution (captured by more SAR operations?) 
%but with vibrations along 3 directions due to the vibration during UAV's flight; 
%(2) R2P can only reconstruct a single object's point cloud while this system can reconstruct a scene's point cloud 
%consists of multiple objects.\\
%To solve this problem, we design two systems. 
The input to each model is a set of $n$ mmWave radar intensity maps or heatmaps collected by an UAV's SAR operation while
hovering at different locations surrounding a scene of interest. 
We will discuss the design details of these two models in the following.

\subsection{Model 1}
\begin{figure}[htb!]
	\centerline{
		\includegraphics[width=0.45\textwidth]{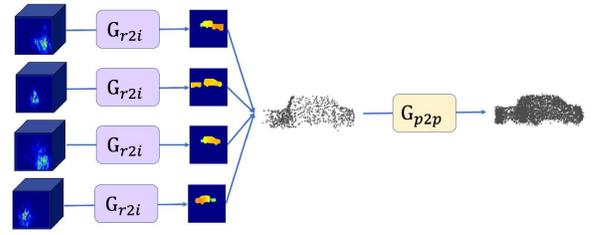}
		
	}
	\caption{Processing pipeline of Model 1. 
		The generated depth images after stage 1 are directly projected into 3D space 
		and form a point cloud containing $m$ objects,
		and then it is fed into $G_{p2p}$ to output an accurate, complete and smooth point cloud of 
		those $m$ objects.
	}
	\label{flow_chart_2}
\end{figure}

Model 1 is our recent work 3DRIMR \cite{sun20213drimr} with R2P \cite{sun2022icann} 
as its second stage generator. 
For completeness we show the model in Fig. \ref{flow_chart_2}.
It consists of two stages. In stage 1, generator network $G_{r2i}$ generates depth images 
based on the radar scans of a scene from multiple view points. Those depth images 
can be directly converted to a rough point cloud. In stage 2, the rough point cloud
is processed by another generator network to produce final dense and smooth point clouds
of the multiple objects in the scene. The training of the two generators are based on conditional 
Generative Adversarial Network (GAN) architecture. 
For details, see \cite{sun20213drimr,sun2022icann}.

\subsection{Model 2}
\begin{figure*}[htb!]
	\centerline{
		\includegraphics[width=0.9\textwidth]{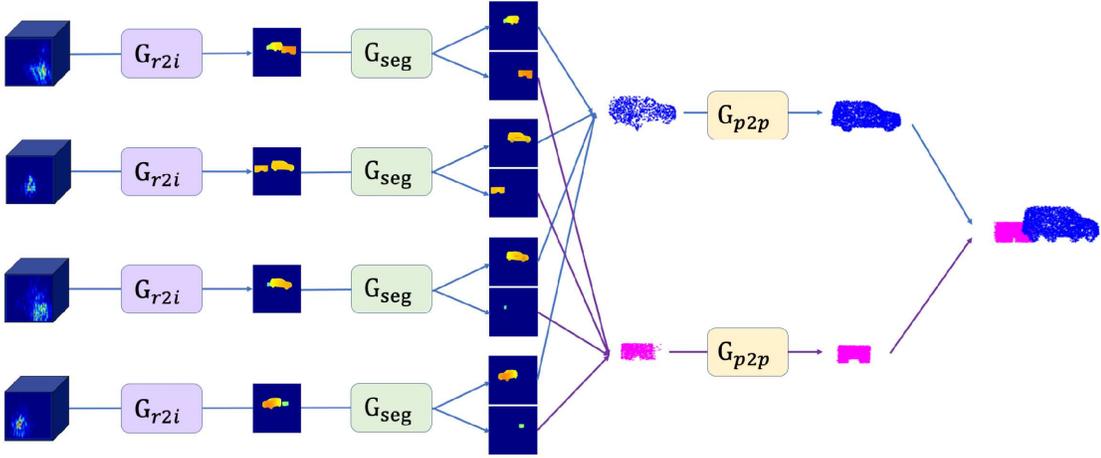}
	}
	\caption{Processing pipeline of Model 2. 
		Each one of $n$ mmWave radar heatmaps (each containing $m$ objects) captured from $n$ view points are fed into $G_{r2i}$,  
		which outputs the corresponding depth image of $m$ objects. 
		%(Note that depth images are all grayscale images, we colormap them here in order to intuitively show the pixel values of them.)
		Then each depth image is processed by an image segmentation network
		% to do semantic segmentation of each pixel
	 which separates the depth image into $m$ depth images and each of them contains only one object.
		Then the $n$ depth images of the same object will be projected into 3D space and form a point cloud, 
		which is then fed into $G_{p2p}$ to output an accurate, complete and smooth point cloud of the object. 
		Finally, all $m$ point clouds of those $m$ objects in the scene will be combined together.
	}
	\label{flow_chart_1}
\end{figure*}

%According to our previous work R2P \cite{sun2022icann},
%reconstructing an object's 3D shape in the format of point clouds is challenging.
%Hence, to reconstruct a 3D scene containing multiple objects is even more challenging.
We add a segmentation stage into Model 1 to from Model 2. 
The idea is to separate the objects in the depth images (obtained from stage 1), and then feed
those individual objects' depth images into the next stage to produce final 3D point clouds
of individual objects, and then finally combine them back into the original space.  
%Since we have shown that R2P can reconstruct one object's 3D shape well,
%we can separate objects in a scene first, and then utilize R2P to reconstruct each object's point clouds, 
%and finally combine the results of those objects.
%Therefore, we designed our 1st system, whose
This model's processing pipeline is illustrated in Fig. \ref{flow_chart_1}.
%We describe each step in the following.
Conditional GAN architecture is also used to train the model. 
\begin{enumerate}
	\item Stage 1: Each one of $n$ mmWave radar intensity map captured from $n$ view points is fed into generator $G_{r2i}$, 
		which can output the corresponding depth image from each viewpoint.
	\item Stage 2: An image segmentation network (e.g., Pix2Pix \cite{isola2017image}) takes those generated depth images as input,
		and then does semantic segmentation of each pixel. 
		Based on the output annotation of each pixel, 
		it can separate a depth image with multiple objects into $m$ depth images, each containing only one object.
	\item Stage 3: For each object, the model projects its corresponding $n$ depth images from $n$ viewpoints into a 3D space
		to form a coarse point cloud of the object. Then for each object, a $G_{p2p}$ network takes its coarse point cloud as input,
		 and outputs an accurate, complete and smoother point cloud of this object. Finally, 
		 the model combines all point clouds of those $m$ objects in the scene 
		 to get the final reconstructed point clouds of multiple objects.	
\end{enumerate}

\subsection{Image Segmentation Network of Model 2}
The segmentation stage of Model 2 
%is a image-based task, our goal 
is to label every pixel of a depth image as either a target object or background. 
%\textit{object $i$} or \textit{background},
We use the popular Pix2Pix \cite{isola2017image} network, 
which is a image-to-image translation conditional generative adversarial network (cGAN), in the segmentation stage. 
The generator is a 2D encoder-decoder convolutional neural network (CNN), 
which can map a grayscale depth image into its annotated image.
The discriminator is a simple 2D CNN, which can output a score to indicate whether the generator's output is good or not.
%The code we used to implemente depth image segmentation here can be found in their GitHub page at \cite{pix2pix_git}.

%\bigskip
\noindent \textbf{Remarks.}
Since 3DRIMR/R2P \cite{sun2022icann} can reconstruct a single object well,
% only if the input coarse point cloud is not so "terrible".
we would like to explore its effectiveness of  
%(i.e., Model 1) 
reconstructing multiple objects, thus we investigate Model 1. 
%Note that if the input coarse point cloud is not so 
On the other hand, since image segmentation may be able to separate multiple objects from each other in a scene and hopefully 
reconstructing single objects might be easier, we come up with the design of Model 2. 
However, introducing the segmentation stage will inevitably add more uncertain errors in the formation of those coarse point clouds,
%(as input to Stage 3 of Model 2),
%which are projected from the depth images after segmentation, 
%If the input coarse point clouds to $G_{p2p}$ (i.e., the outputs of the first stage of Models 1 or 2) lose too much shape characteristics,
and then those additional errors may make $G_{p2p}$ not be able to extract useful features and fail to reconstruct 
the 3D shapes of objects.
For example, if some pixels of a small-sized object are annotated as background pixels during the segmentation stage, 
then the generated coarse point cloud may totally lose the shape characteristics of the object.
%From this concern, we remove segmentation process  from the 1st system's pipeline,
%and just project the generated depth image containing multiple obejcts into 3D space,
%and then directly pass it into R2P's $G_{p2p}$ to see if it can also reconstruct multiple point clouds from different objects at the same time.

% ================================================================================================================
\section{EXPERIMENTS}\label{sec_imp}

\subsection{Datasets}

\begin{figure}[htb!]
	%\begin{figure}[!h]
	\centering
	\includegraphics[width=2in]{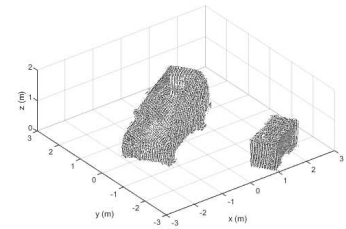}
	\caption{Example scene of two objects.}
	%\label{example_reward}
	%\end{figure}
	\label{fig_example_scene}
\end{figure}

\begin{figure}[htb!]
	%\begin{wrapfigure}{R}{0.40\textwidth}%{60mm}
	\centerline{
		\begin{minipage}{1.75in}
			\begin{center}
				\setlength{\epsfxsize}{1.75in}
				%\epsffile{./images/encoding_network.eps}\\
				%\epsffile{./figures/no_vib_car_1_placement_1_n_combined_3view_Sum.eps}\\
							\epsffile{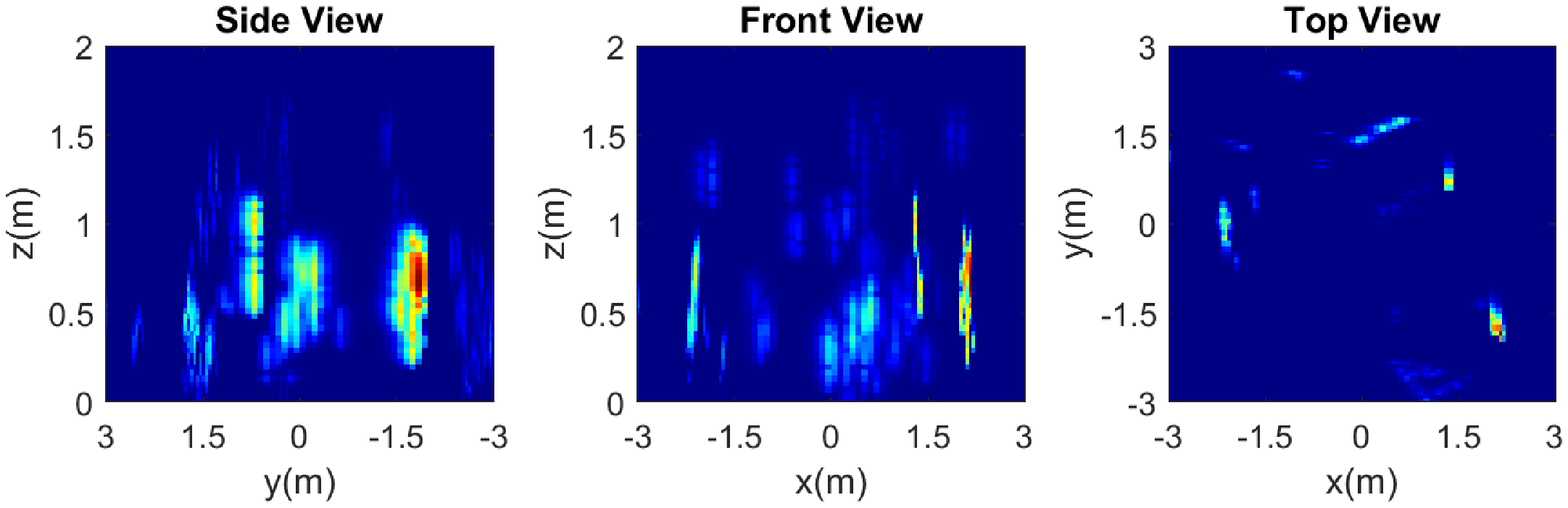}\\
				\small{(a) Radar energy from normal SAR operation.}\normalfont 
			\end{center}
		\end{minipage}
		%\hspace{2mm}
		\begin{minipage}{1.3in}
			\begin{center}
				\setlength{\epsfxsize}{1.3in}
				%\epsffile{./images/encoding_job.eps}\\
				%\epsffile{./figures/no_vib_car_1_placement_1_n_combined_heatmap3d_Sum_preview.eps}\\
					\epsffile{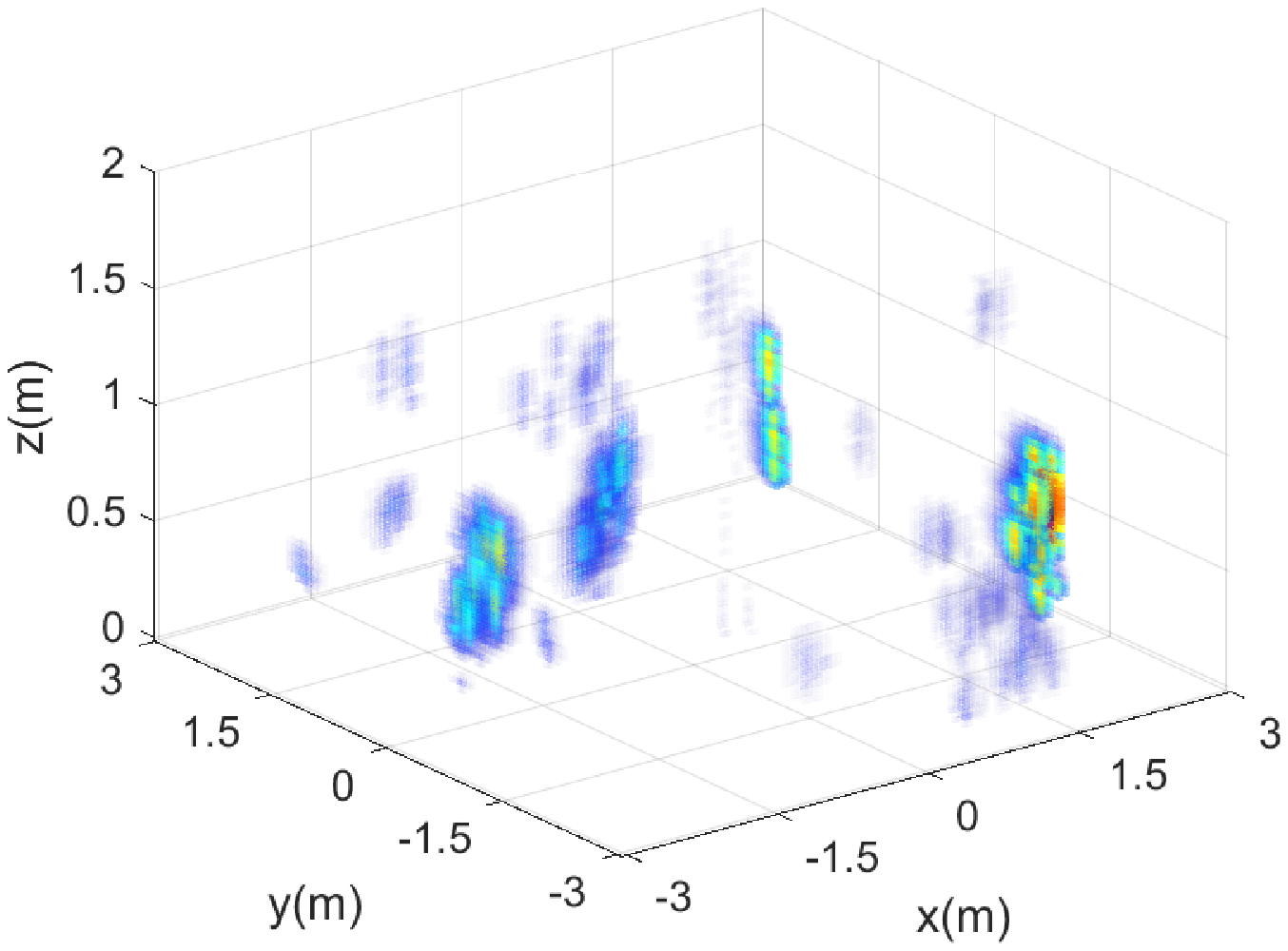}\\
					\small{(b) FFT heatmap from normal SAR.}\normalfont
			\end{center}
		\end{minipage} 
	}
	\centerline{
		\begin{minipage}{1.75in}
			\begin{center}
				\setlength{\epsfxsize}{1.75in}
				%\epsffile{./figures/vib_car_1_placement_1_v_combined_3view_Sum.eps}\\
				\epsffile{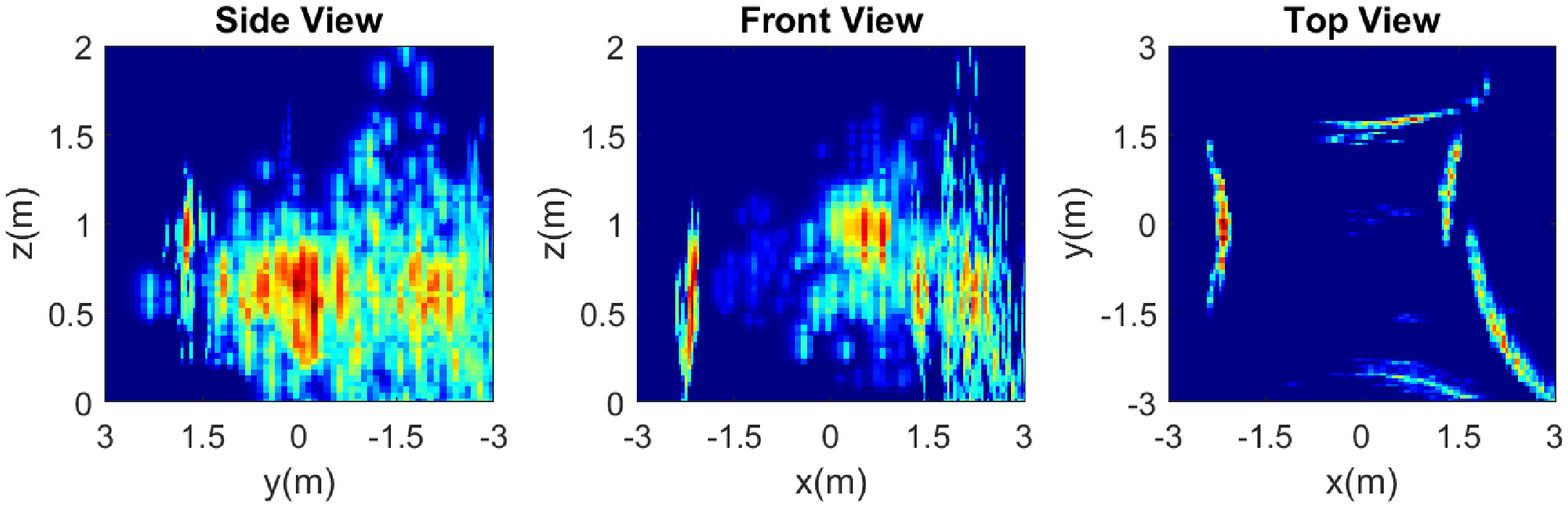}\\
				\small{(c) Radar energy from vibrating UAV's SAR.}\normalfont
			\end{center}
		\end{minipage}
		\begin{minipage}{1.3in}
		\begin{center}
			\setlength{\epsfxsize}{1.3in}
			%\epsffile{./figures/vib_car_1_placement_1_v_combined_heatmap3d_Sum_preview.eps}\\
			\epsffile{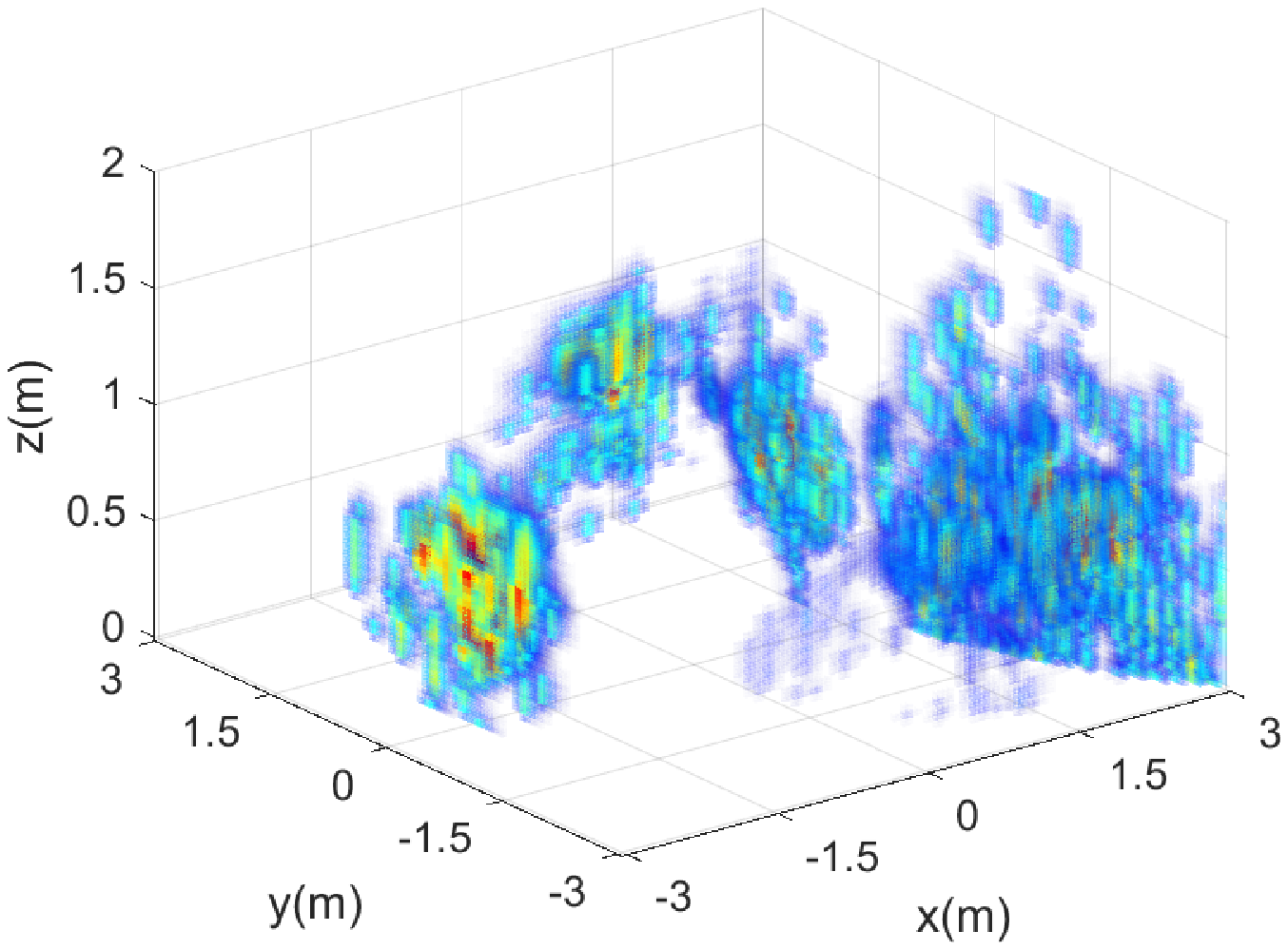}\\
			\small{(d) FFT heatmap from vibrating UAV's SAR. }\normalfont
		\end{center}
	\end{minipage}
	}
	\caption{Comparison between normal SAR and the SAR of vibrating UAV. }
	\label{fig_compare_SARs}
	%\end{wrapfigure}
\end{figure}

Our experiments are mainly based on synthetic datasets, as collecting data via our current UAV platform 
is very time consuming. We first use OptiTrack \cite{optitrack} system to measure the deviation distances of 
a hovering UAV along x-, y-, and z-axis from its 
intended stable hovering position for SAR operation (scanning a space), 
and then based on the collected statistics we add noise (caused by hovering vibration) 
into the synthetic data generating process \cite{HawkEye, sun20213drimr}. 
%generate synthetic raw radar energy signals of the space.
We next show via an example that the radar data collected by a vibrating UAV's SAR operation visually looks
quite different from the data of a normal stable SAR operation. 
Fig. \ref{fig_example_scene} shows a scene where a car and a desk are placed, and 
Fig. \ref{fig_compare_SARs} shows the analysis of the radar data of the scene. 
As shown in Fig. \ref{fig_compare_SARs}, a normal SAR operation gives cleaner 
and more distinctly clustered radar energy maps and FFT heatmaps (Fig. \ref{fig_compare_SARs} (a) and (b)),
when compared with the maps obtained from a vibrating UAV's SAR operation. Our experiments reported in this paper are all based on the vibrating UAV's SAR data. 

In addition, we follow a procedure that is similar to 3DRIMR \cite{sun20213drimr} 
to generate ground truth depth images and point clouds.
To generate the ground truth annotated images that are used to train the segmentation network,
we manually set the background pixels to \textit{black} color, and the pixels of 
the same object to the same RGB color (except black color).

\subsection{Model Training and Testing}

The training and testing processes of $G_{r2i}$ and $G_{p2p}$ in Model 1 are the same as those in \cite{sun2022icann} 
except that in this paper both the training and testing data are point clouds of multiple objects rather than individual ones.
Similarly, the training processes of $G_{r2i}$ and $G_{p2p}$ in Model 2 are the same as those in \cite{sun2022icann}.
As for the segmentation stage of Model 2, 
we use $2400 \times 4$ views, totally $9600$ pairs of images to train $G_{seg}$ for $200$ epochs,
and the learning rate is $2 \times 10^{-4}$ for the first $10$0 epochs and linearly reduced to $0$ for the rest $100$ epochs.

\subsection{Evaluation Results}

In our experiments, $G_{p2p}$ takes an input point cloud with 1024 points and generates an output point cloud 
with 4096 points.
If we combine $m$ objects' generated point clouds of Model 2, the final output point clouds will contain $4096 \times m$ points,
which is $m$ times denser than the output point clouds of Model 1.
%This is an advantages of Model 2 that Model 2 can generate denser output point clouds than Model 1.
For the sake of fairness, when comparing the
Chamfer Distance (CD) and Earth Mover’s Distance (EMD) of both models, 
we randomly sample 4096 points from the output point clouds of Model 2 
so that the output point clouds using these two methods have the same number of points.
Then we calculate the two models' CD/EMD with the ground truth point clouds containing 4096 points respectively.
%The quantitative results are shown in Table \ref{loss_table}.

% table for different settings
\renewcommand{\arraystretch}{1.5} 
\begin{table}
	\centering
	\fontsize{6.5}{8}\scriptsize
	\begin{tabular}{|l|l|cc|cc|}
		\hline
		\multicolumn{1}{|c|}{\multirow{2}{*}{Scene}} & \multicolumn{1}{c|}{\multirow{2}{*}{Method}} & \multicolumn{2}{c|}{CD}          & \multicolumn{2}{c|}{EMD}         \\ \cline{3-6} 
		\multicolumn{1}{|c|}{}                       & \multicolumn{1}{c|}{}                        & \multicolumn{1}{c|}{avg.} & std. & \multicolumn{1}{c|}{avg.} & std. \\ 
		\hline \hline
		\multirow{4}{*}{2 objects}                   
		& Model 1 (CD)                              & \multicolumn{1}{c|}{0.2}  & 0.06 & \multicolumn{1}{c|}{2.74} & 0.8  \\ \cline{2-6} 
		& Model 2 (CD)                              & \multicolumn{1}{c|}{0.21} & 0.06 & \multicolumn{1}{c|}{4.74} & 1.04 \\ \cline{2-6} \cline{3-6} 
		& Model 1 (EMD)                             & \multicolumn{1}{c|}{0.22} & 0.05 & \multicolumn{1}{c|}{0.57} & 0.32 \\ \cline{2-6} 
		& Model 2 (EMD)                             & \multicolumn{1}{c|}{0.25} & 0.07 & \multicolumn{1}{c|}{3.79} & 0.87 \\ 
		\hline \hline
		\multirow{4}{*}{3 objects}                   
		& Model 1 (CD)                              & \multicolumn{1}{c|}{0.36} & 0.12 & \multicolumn{1}{c|}{4.58} & 0.63 \\ \cline{2-6} 
		& Model 2 (CD)                              & \multicolumn{1}{c|}{0.3}  & 0.08 & \multicolumn{1}{c|}{5.81} & 1.28 \\ \cline{2-6} \cline{3-6} 
		& Model 1 (EMD)                             & \multicolumn{1}{c|}{0.31} & 0.08 & \multicolumn{1}{c|}{0.74} & 0.28 \\ \cline{2-6} 
		& Model 2 (EMD)                             & \multicolumn{1}{c|}{0.33} & 0.1  & \multicolumn{1}{c|}{4.46} & 1.15 \\ \hline
	\end{tabular}
	\caption {Quantitative Results under different settings. 
		We conduct our experiments on 2 different scenes: 
		a scene containing 2 objects (a car and a desk) and a scene containing 3 objects (a car, a desk and a robot arm).
		 The loss type CD or EMD in the parenthesis of method names indicates the distance functions 
		 %$d_1$ and $d_2$ 
		 used in training. For example, Model 1 (CD) means the Model 1 is used and CD is used as the distance functions 
		 %of both $d_1$ and $d_2$ 
		 during training.}
	\label{loss_table}
\end{table}

Fig. \ref{pc_plots} shows the visual comparison of the two models in reconstructing two objects. We see that
both models can give reasonably good reconstruction performance visually. However, there might be extra
points between two objects in the point clouds generated by Model 1. 
This is due to the fact that Model 1 attempts to reconstruct multiple objects together.
We also see that the point clouds generated by Model 2 is smoother and denser. 
We have similar observations of the experiments with three objects.
Fig. \ref{pc_plots} also shows that both models are fairly robust to the highly 
distorted/noisy SAR data (Fig. \ref{fig_compare_SARs}) caused by unstable UAV hovering. 

Table \ref{loss_table} shows the test results of  both Model 1 and Model 2 
on two different scenes: one with two objects and the other one with three objects.
We can see that the two models' test results of CDs are almost the same for the two scenes.
However their performance are different in terms of EMD.
For the 2-object scene, if the loss functions used in training
%in R2P 
are CDs, then the average test result EMD between generated point clouds and ground truth point clouds 
of Model 2 is around 1.7 times of Model 1.
If the loss functions in training are EMDs, then the average test result EMD of Model 2 is more than $5.6$ times larger than Model 1.
The scene with 3 objects also shows the similar results.

\begin{figure}[htb!]
	%\begin{wrapfigure}{R}{0.40\textwidth}%{60mm}
	\centerline{
		\begin{minipage}{1.5in}
			\begin{center}
				\setlength{\epsfxsize}{1.5in}
				%\epsffile{./images/encoding_network.eps}\\
				\epsffile{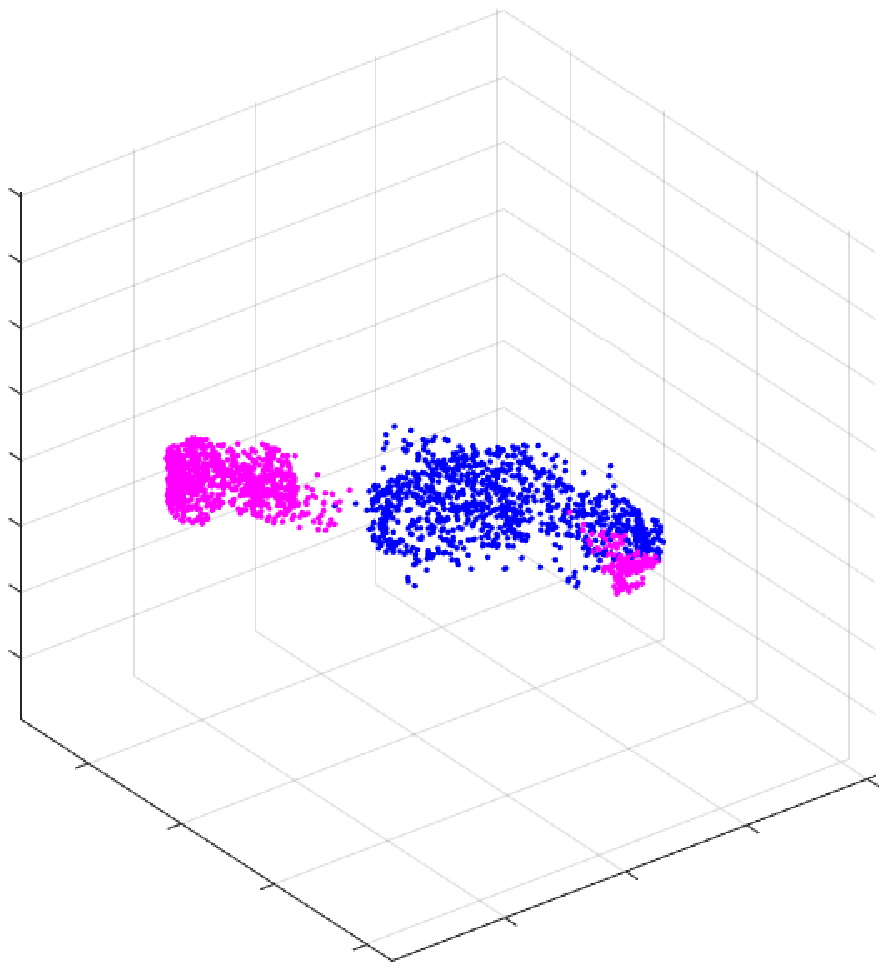}\\
				\small{(a) Coarse Point Cloud.}\normalfont
			\end{center}
		\end{minipage}
		%\hspace{2mm}
		\begin{minipage}{1.5in}
			\begin{center}
				\setlength{\epsfxsize}{1.5in}
				%\epsffile{./images/encoding_job.eps}\\
				\epsffile{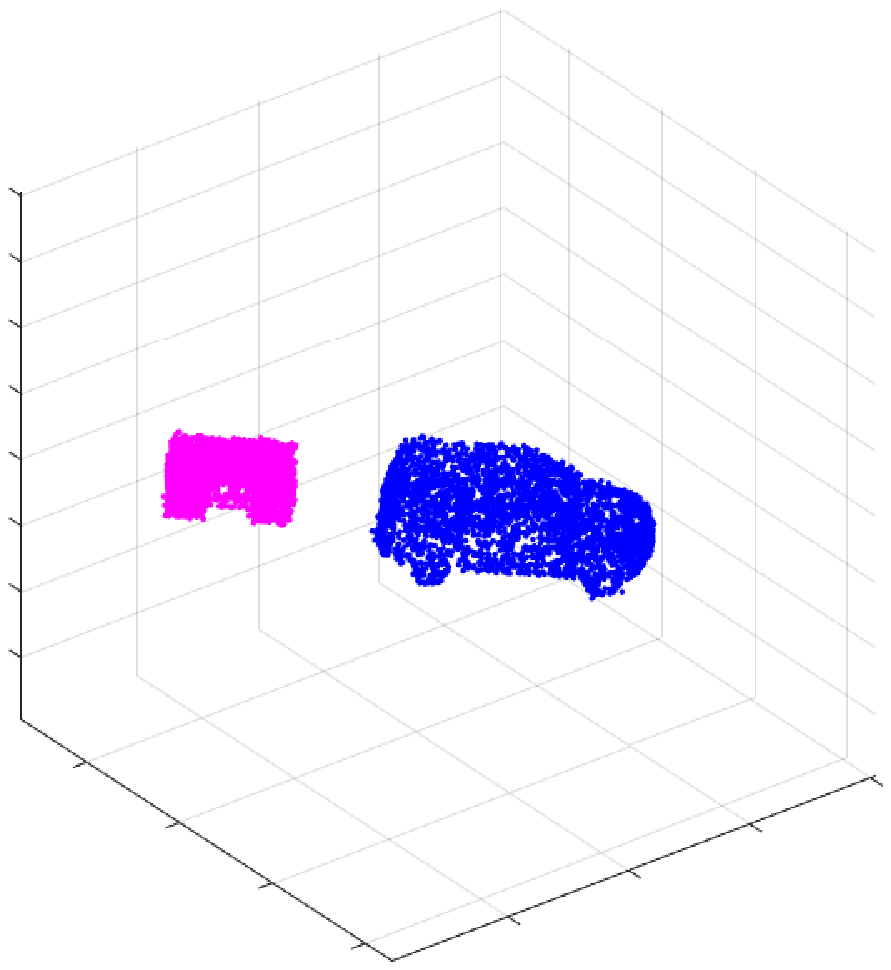}\\
				\small{(b) Ground Truth Point Cloud.}\normalfont
			\end{center}
		\end{minipage} 
	}
	\centerline{
		\begin{minipage}{1.5in}
			\begin{center}
				\setlength{\epsfxsize}{1.5in}
				\epsffile{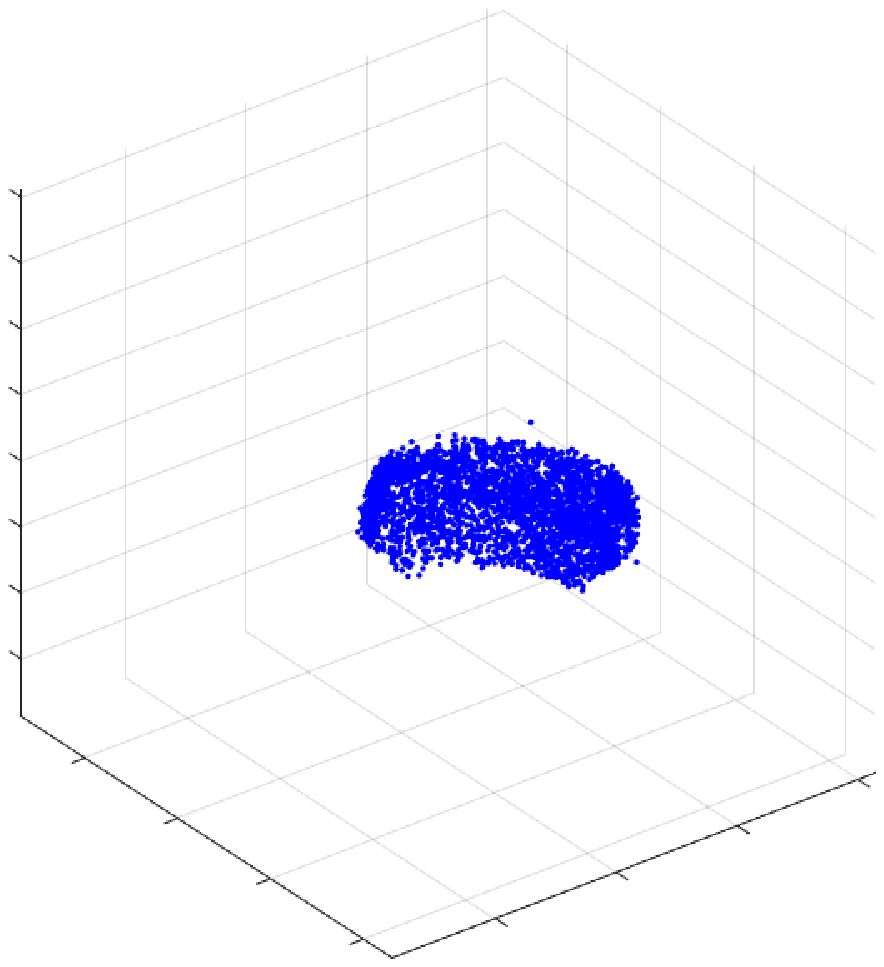}\\
				\small{(c) Output Point Cloud of Model 2 (CF).}\normalfont
			\end{center}
		\end{minipage}
		\begin{minipage}{1.5in}
			\begin{center}
				\setlength{\epsfxsize}{1.5in}
				\epsffile{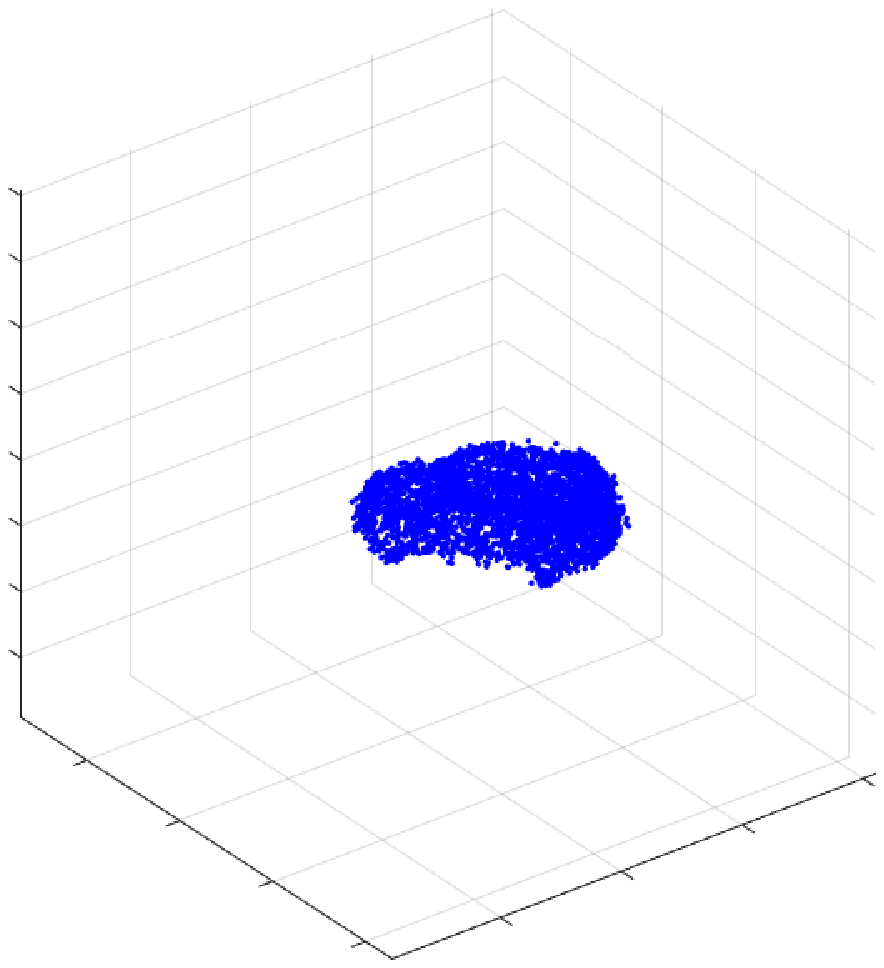}\\
				\small{(d) Output Point Cloud of Model 2 (EMD).}\normalfont
			\end{center}
		\end{minipage}
	}
	\caption{A failure example of Model 2. }
	\label{fig_failure_pc}
\end{figure}

The superior performance of Model 1 over Model 2 in terms of CD/EMD can be explained as follows. 
During the training of $G_{p2p}$ in Model 1, the goal is to reduce the losses (CD/EMD) of all $m$ objects as a whole
between the generated point clouds and the ground truth point clouds.
On the other hand, the $G_{p2p}$ of each object in Model 2 is trained separately, so 
the network is only trained to reduce the losses (CD/EMD) of a single object.
Therefore Model 1 will have much lower overall losses than Model 2 during test or inference. 
Furthermore, we notice that 
the segmentation network in Model 2 can introduce additional reconstruction errors. 
One example is shown in Fig. \ref{fig_failure_pc}. 
Due to poor performance of $G_{seg}$,
%if its input depth image has low quality, 
the projected point cloud of the $m$ objects will contain lots of errors (Fig. \ref{fig_failure_pc}(a)), 
which further causes $G_{p2p}$'s failure.
As shown in Fig. \ref{fig_failure_pc}(c)-(d), the object with too many wrong points (desk in this example) is totally missing
in the output point clouds, and the other object 
%with too many missing points will be 
is wrongly reconstructed 
(e.g., the orientation of the reconstructed car is inversed).

However, Model 2 can generate denser point clouds than Model 1. 
Besides, since Model 2 reconstructs each object separately, there will be no extra points between objects
in the final point cloud,
a problem associated with Model 1. For example, as we can see in Fig. \ref{pc_plots}, the output point clouds of Model 1 (EMD) have some extra points between two objects.

\begin{table*}[h]
	\small
	\centering
	\begin{tabular}{l c c c c }
		% row: 2nd System (CD) ---------------------------------------------------------------------------------------
		Model 1 (CD)
		&
		\begin{minipage}[b]{0.4\columnwidth}
			\centering
			\raisebox{-.5\height}{\includegraphics[width=\linewidth]{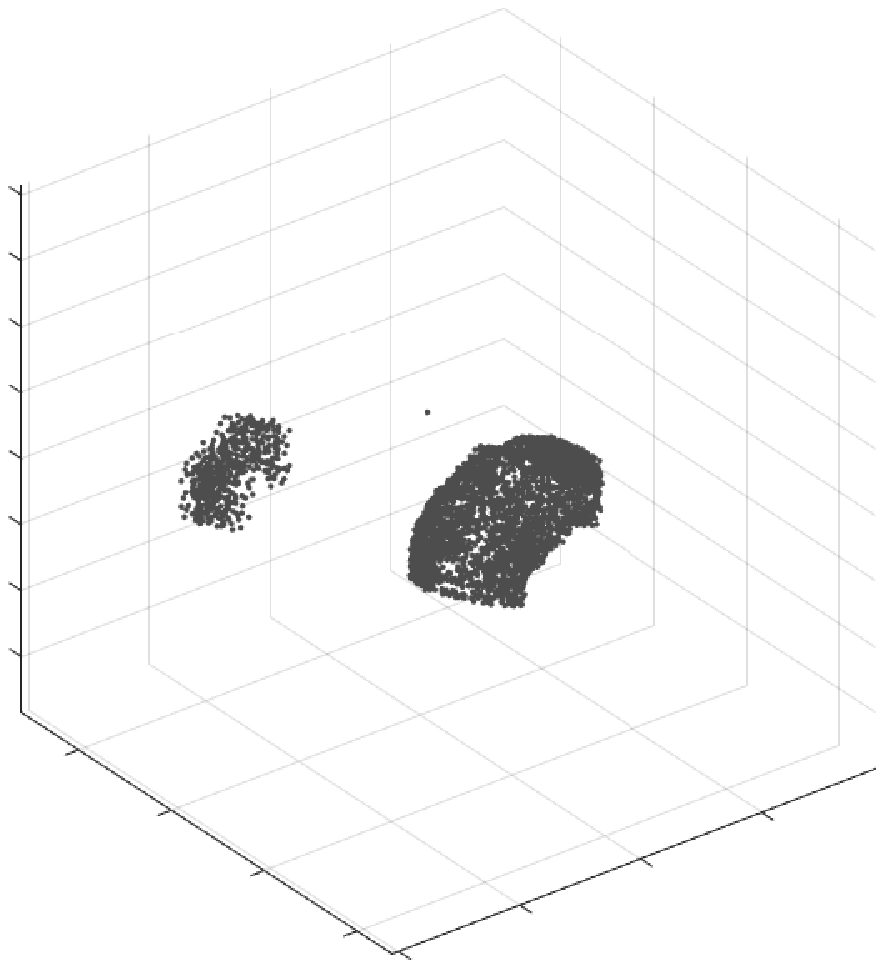}}
		\end{minipage}
		&
		\begin{minipage}[b]{0.4\columnwidth}
			\centering
			\raisebox{-.5\height}{\includegraphics[width=\linewidth]{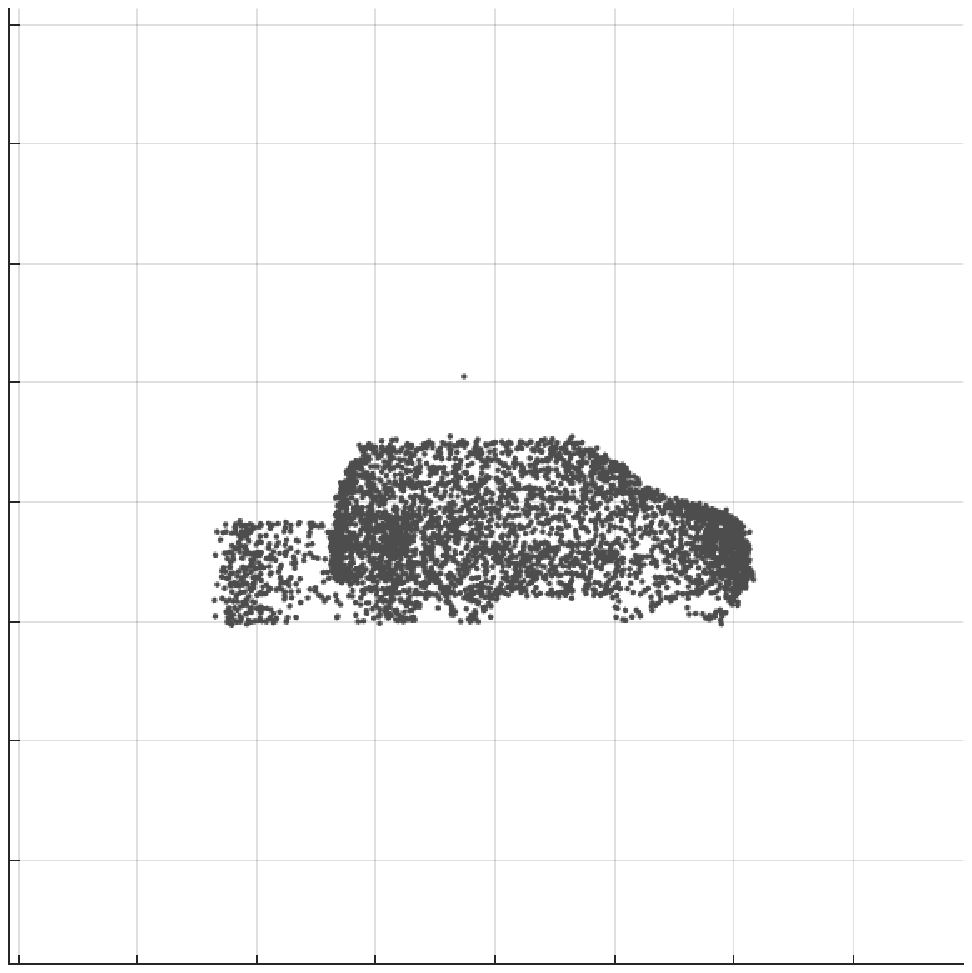}}
		\end{minipage}
		&
		\begin{minipage}[b]{0.4\columnwidth}
			\centering
			\raisebox{-.5\height}{\includegraphics[width=\linewidth]{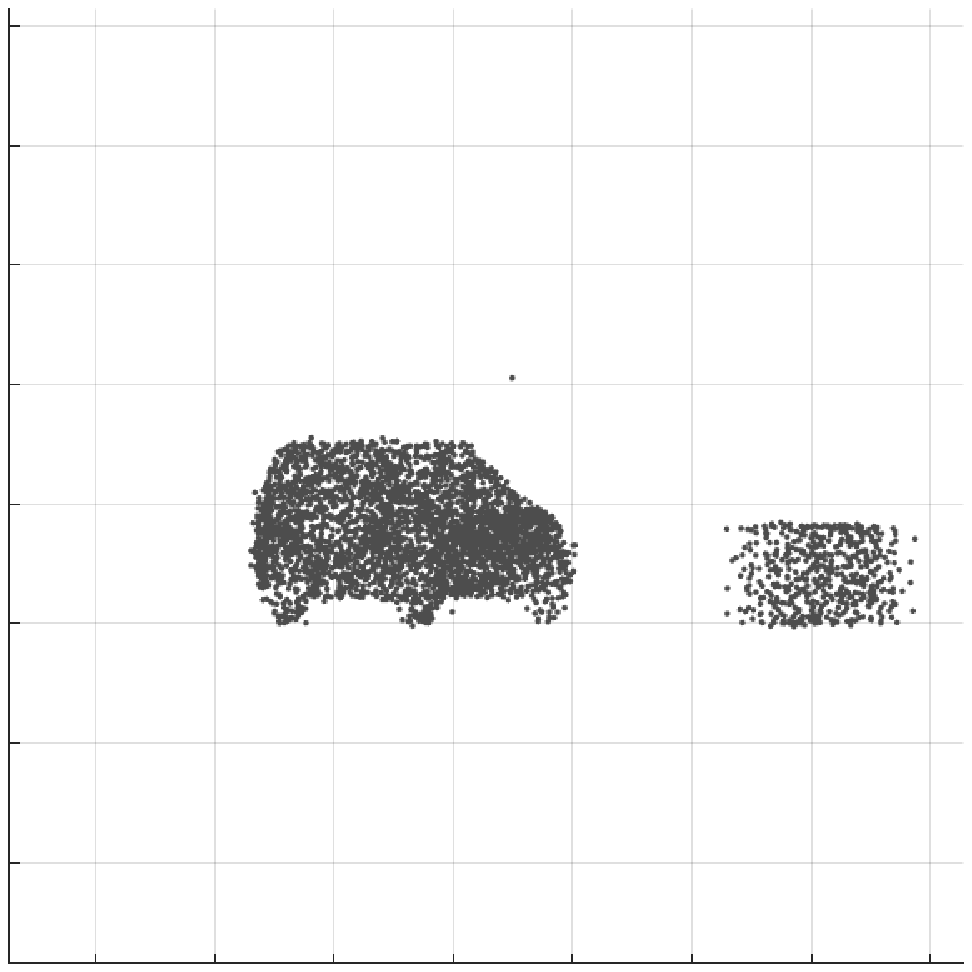}}
		\end{minipage}
		&
		\begin{minipage}[b]{0.4\columnwidth}
			\centering
			\raisebox{-.5\height}{\includegraphics[width=\linewidth]{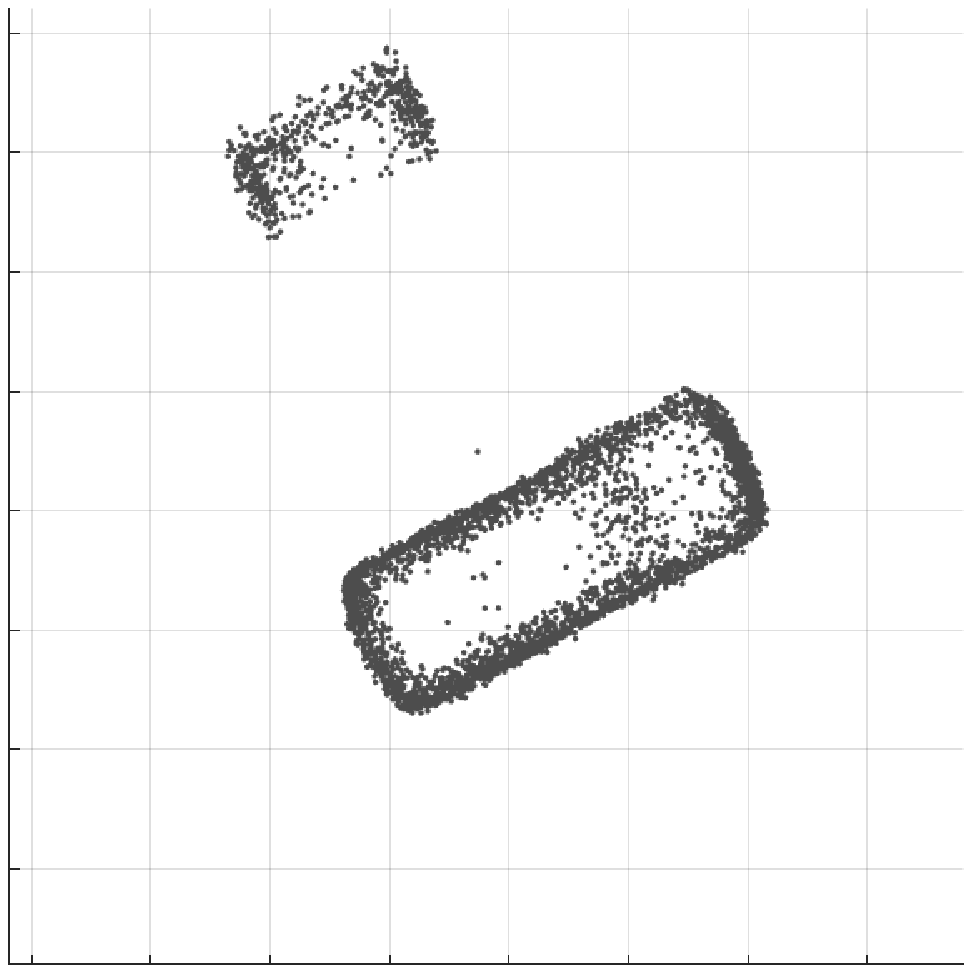}}
		\end{minipage}
		\\
		% row: 2nd System (EMD) ---------------------------------------------------------------------------------------
		Model 1 (EMD)
		&
		\begin{minipage}[b]{0.4\columnwidth}
			\centering
			\raisebox{-.5\height}{\includegraphics[width=\linewidth]{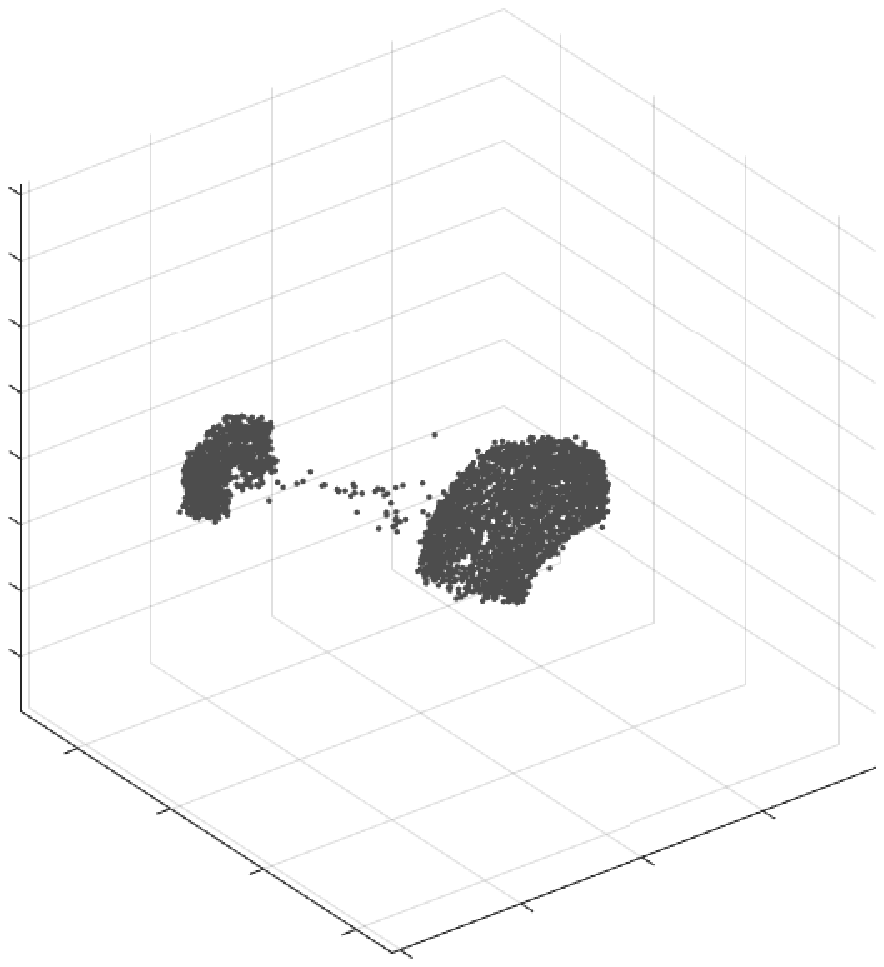}}
		\end{minipage}
		&
		\begin{minipage}[b]{0.4\columnwidth}
			\centering
			\raisebox{-.5\height}{\includegraphics[width=\linewidth]{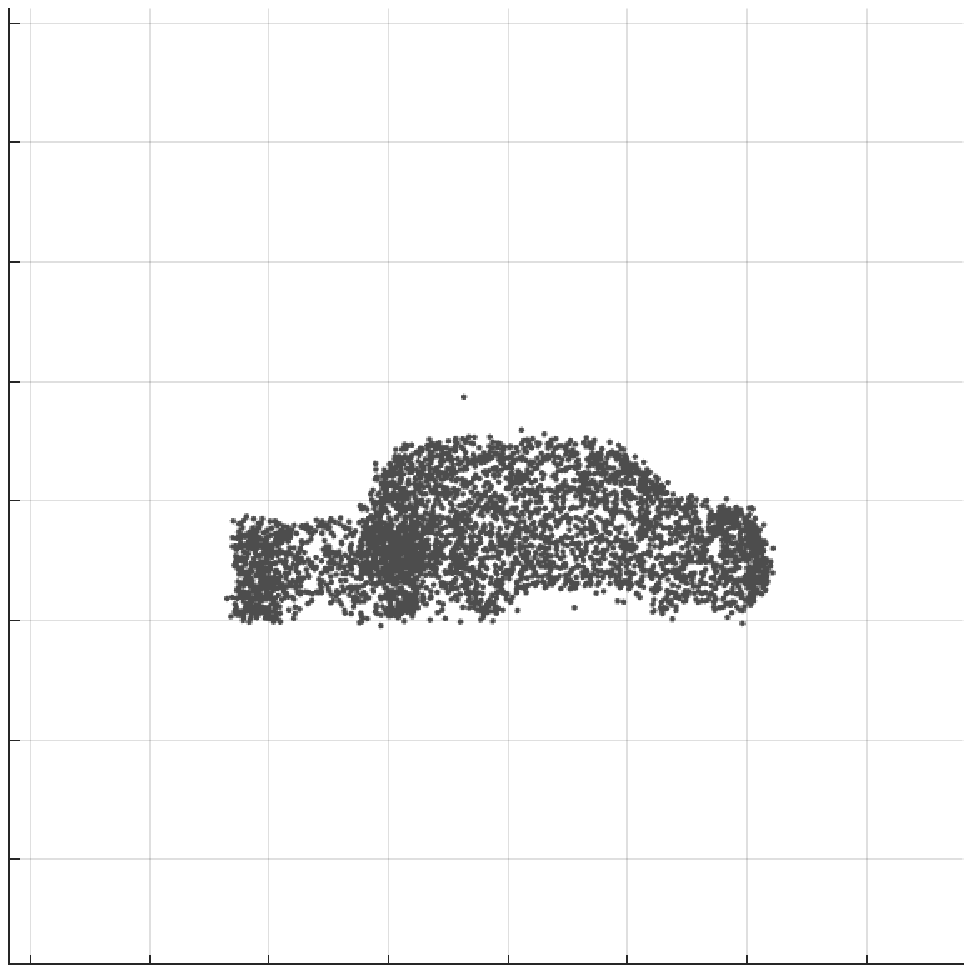}}
		\end{minipage}
		&
		\begin{minipage}[b]{0.4\columnwidth}
			\centering
			\raisebox{-.5\height}{\includegraphics[width=\linewidth]{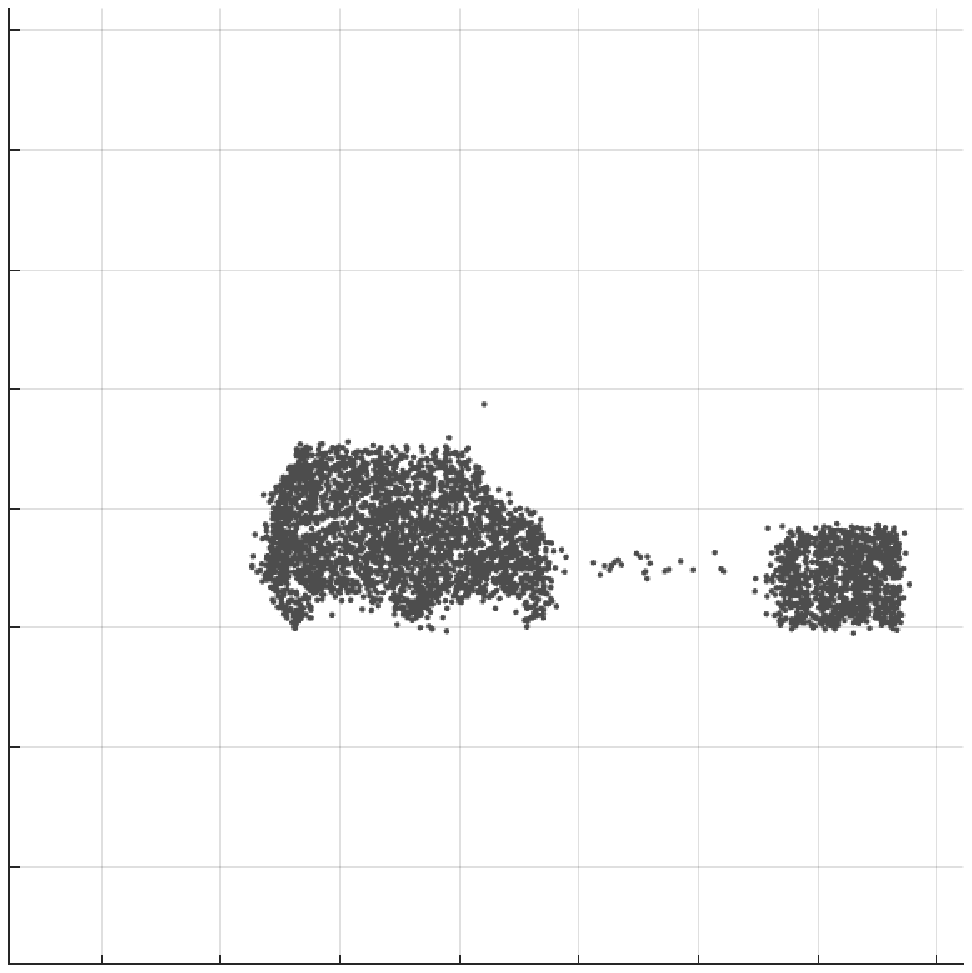}}
		\end{minipage}
		&
		\begin{minipage}[b]{0.4\columnwidth}
			\centering
			\raisebox{-.5\height}{\includegraphics[width=\linewidth]{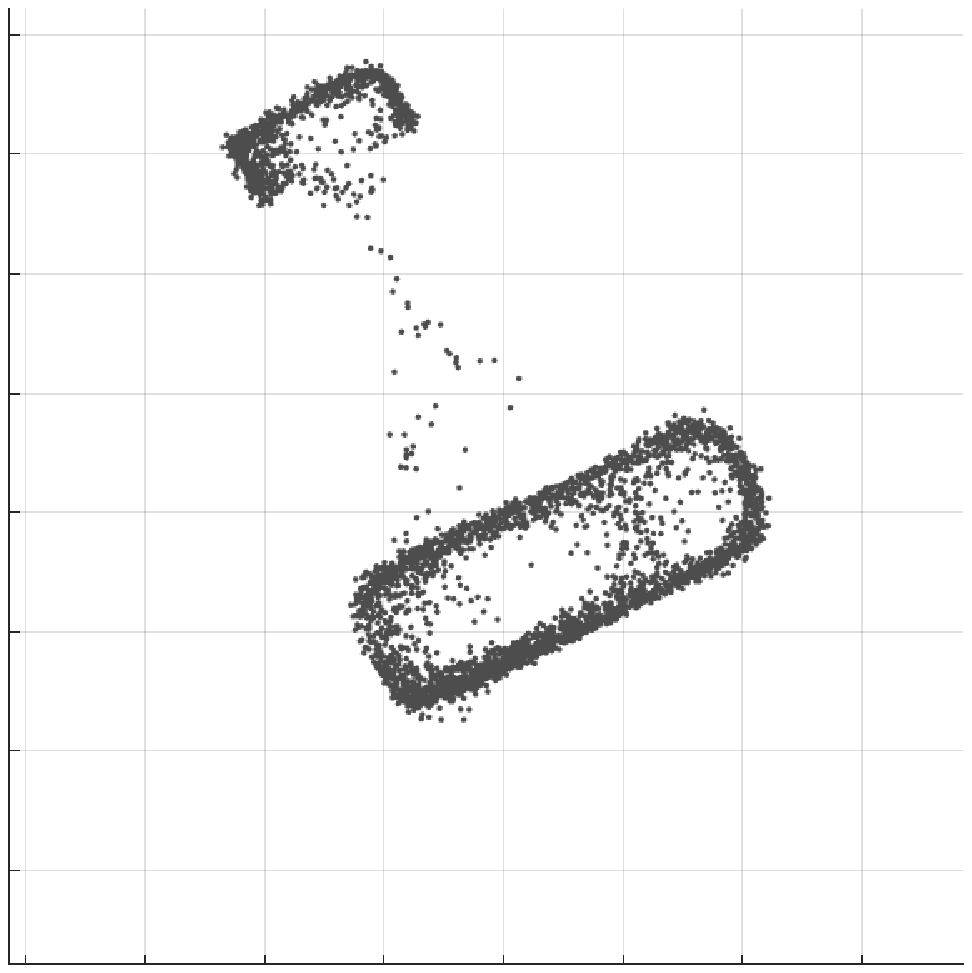}}
		\end{minipage}
		\\
		% row: 1st System (CD) ---------------------------------------------------------------------------------------
		Model 2 (CD)
		&
		\begin{minipage}[b]{0.4\columnwidth}
			\centering
			\raisebox{-.5\height}{\includegraphics[width=\linewidth]{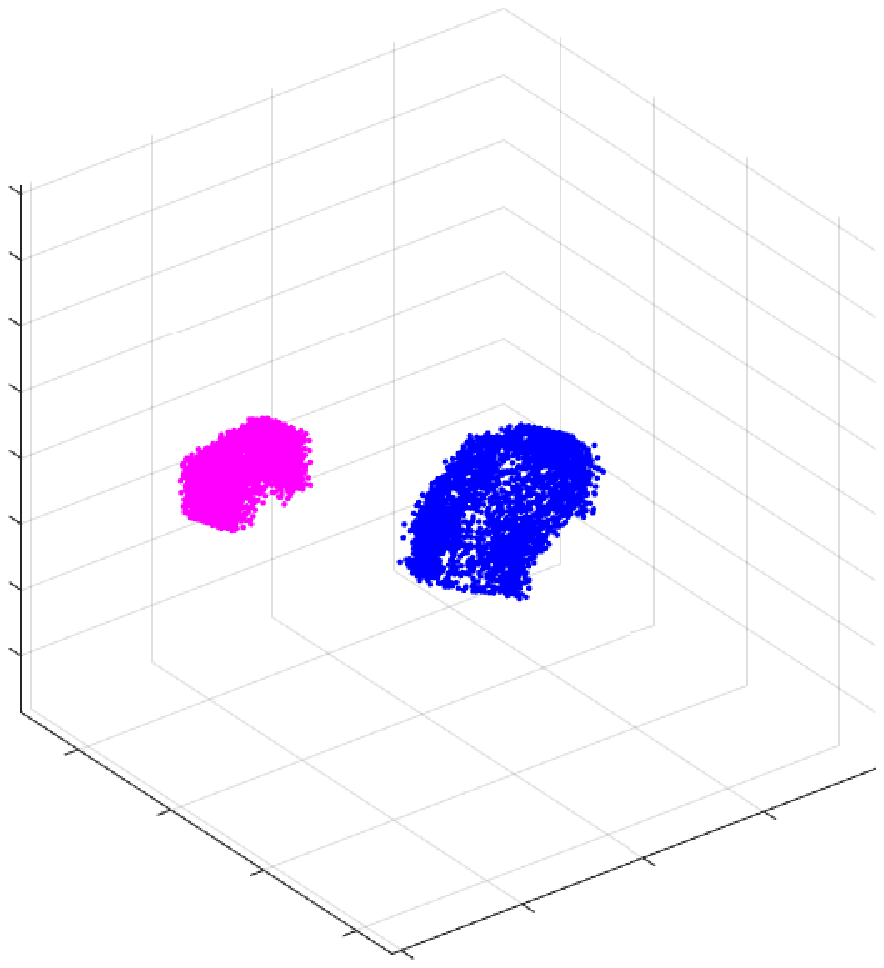}}
		\end{minipage}
		&
		\begin{minipage}[b]{0.4\columnwidth}
			\centering
			\raisebox{-.5\height}{\includegraphics[width=\linewidth]{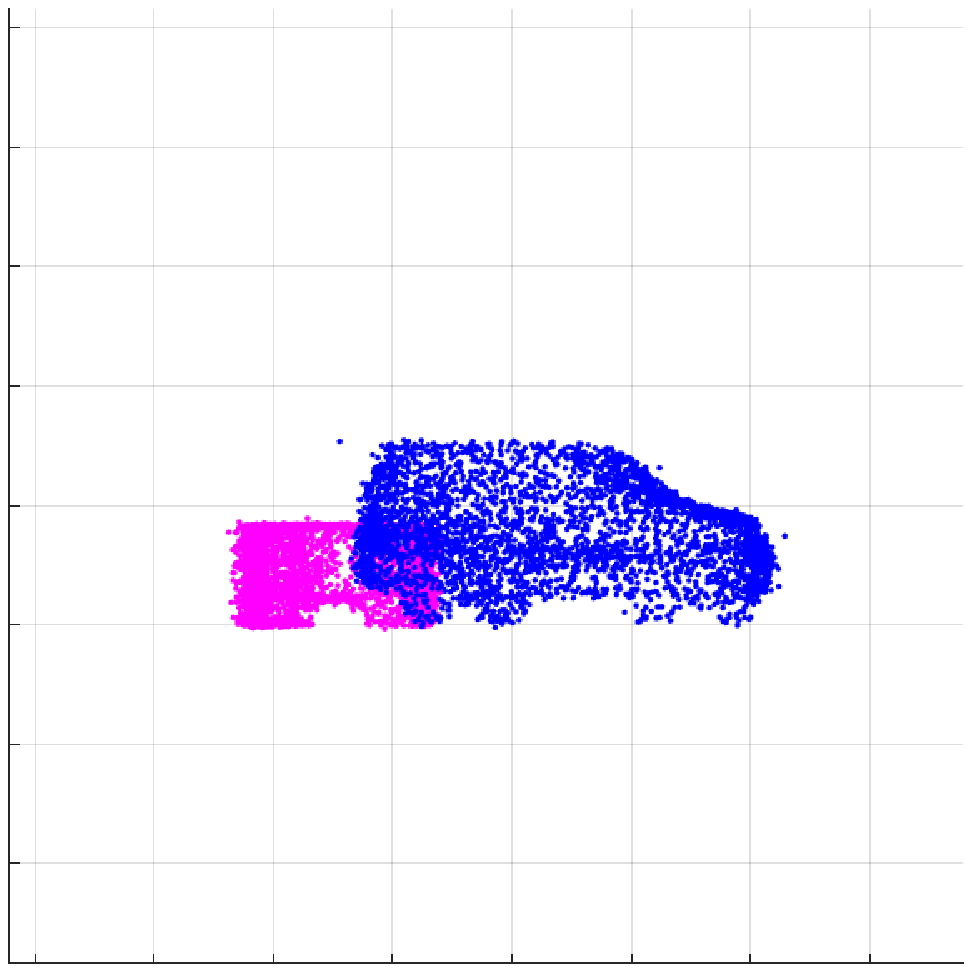}}
		\end{minipage}
		&
		\begin{minipage}[b]{0.4\columnwidth}
			\centering
			\raisebox{-.5\height}{\includegraphics[width=\linewidth]{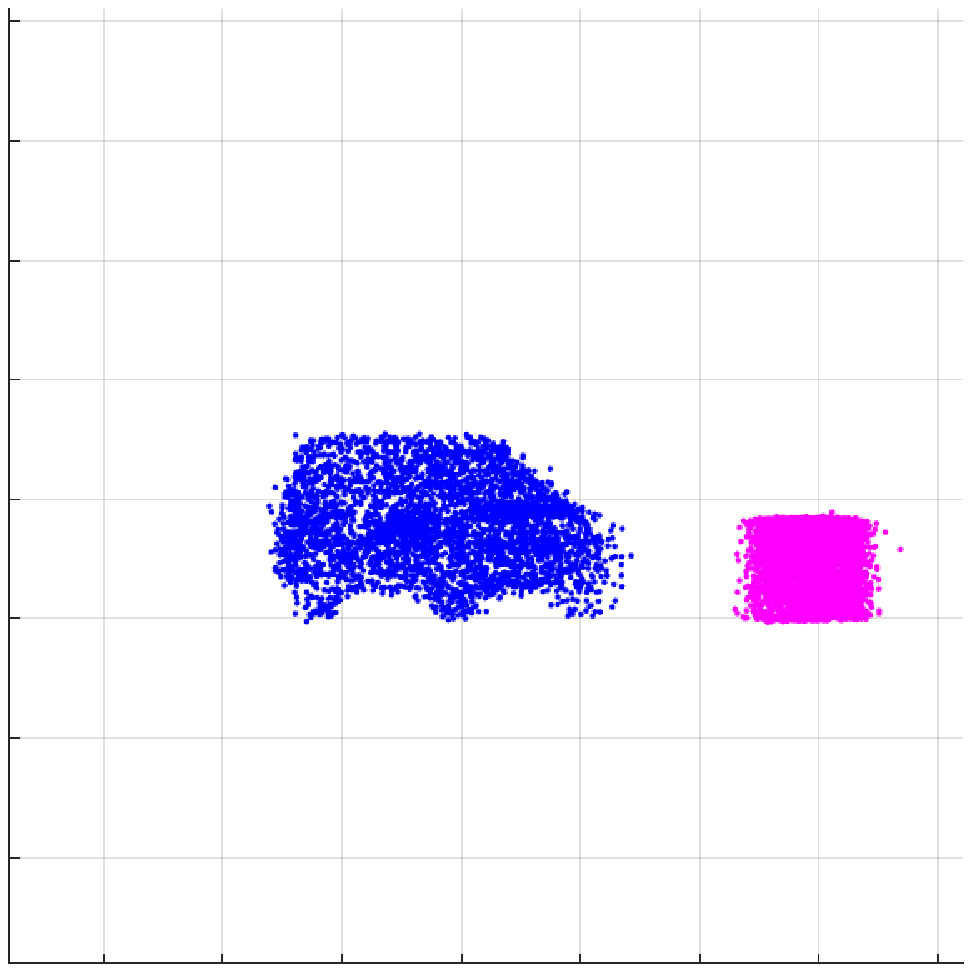}}
		\end{minipage}
		&
		\begin{minipage}[b]{0.4\columnwidth}
			\centering
			\raisebox{-.5\height}{\includegraphics[width=\linewidth]{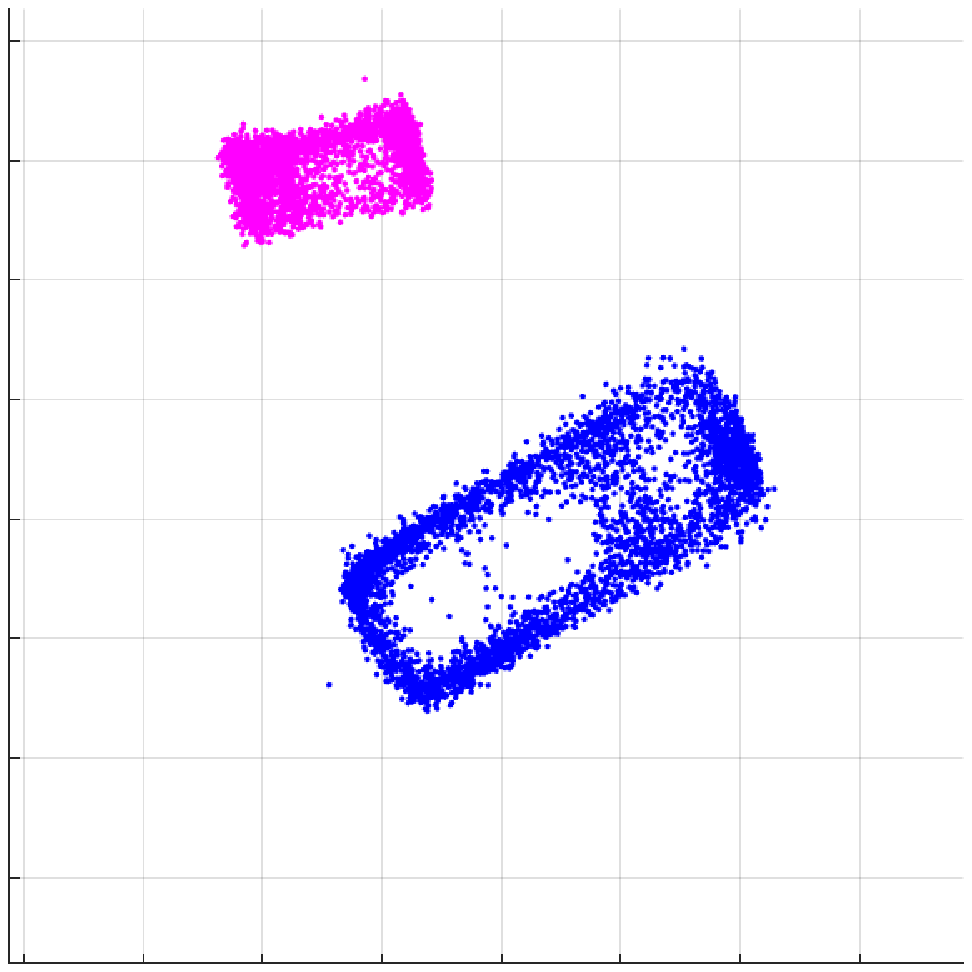}}
		\end{minipage}
		\\
		% row: 1st System (EMD) ---------------------------------------------------------------------------------------
		Model 2 (EMD)
		&
		\begin{minipage}[b]{0.4\columnwidth}
			\centering
			\raisebox{-.5\height}{\includegraphics[width=\linewidth]{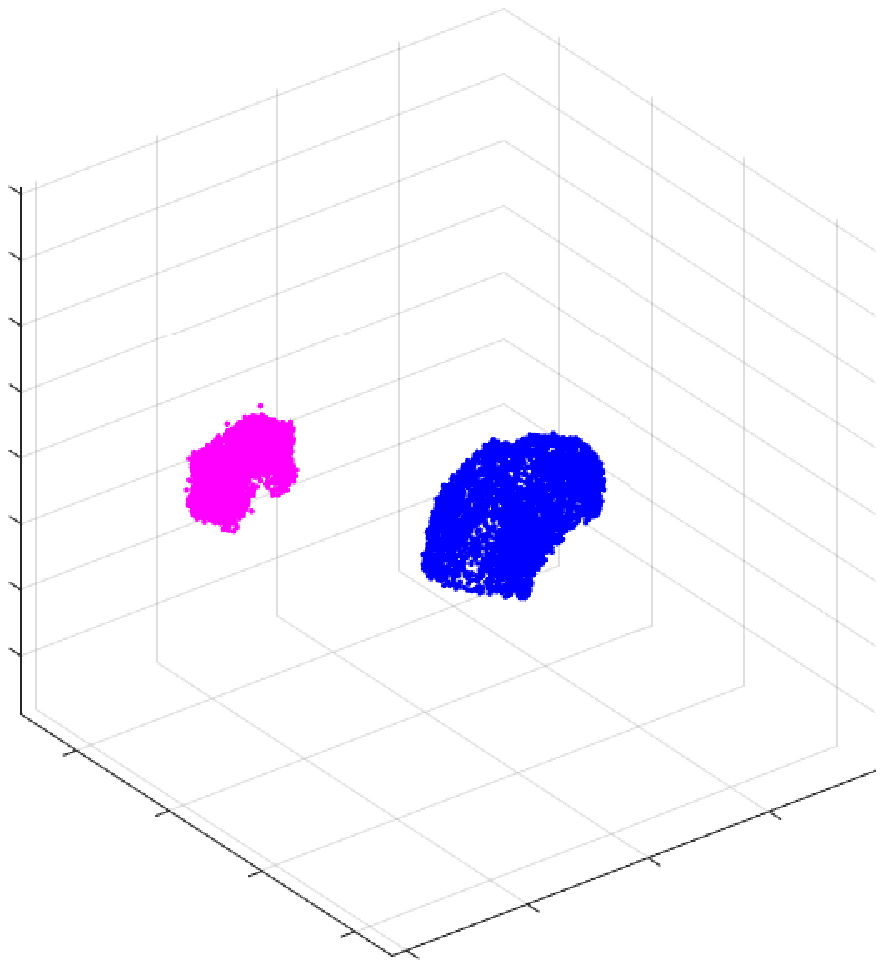}}
		\end{minipage}
		&
		\begin{minipage}[b]{0.4\columnwidth}
			\centering
			\raisebox{-.5\height}{\includegraphics[width=\linewidth]{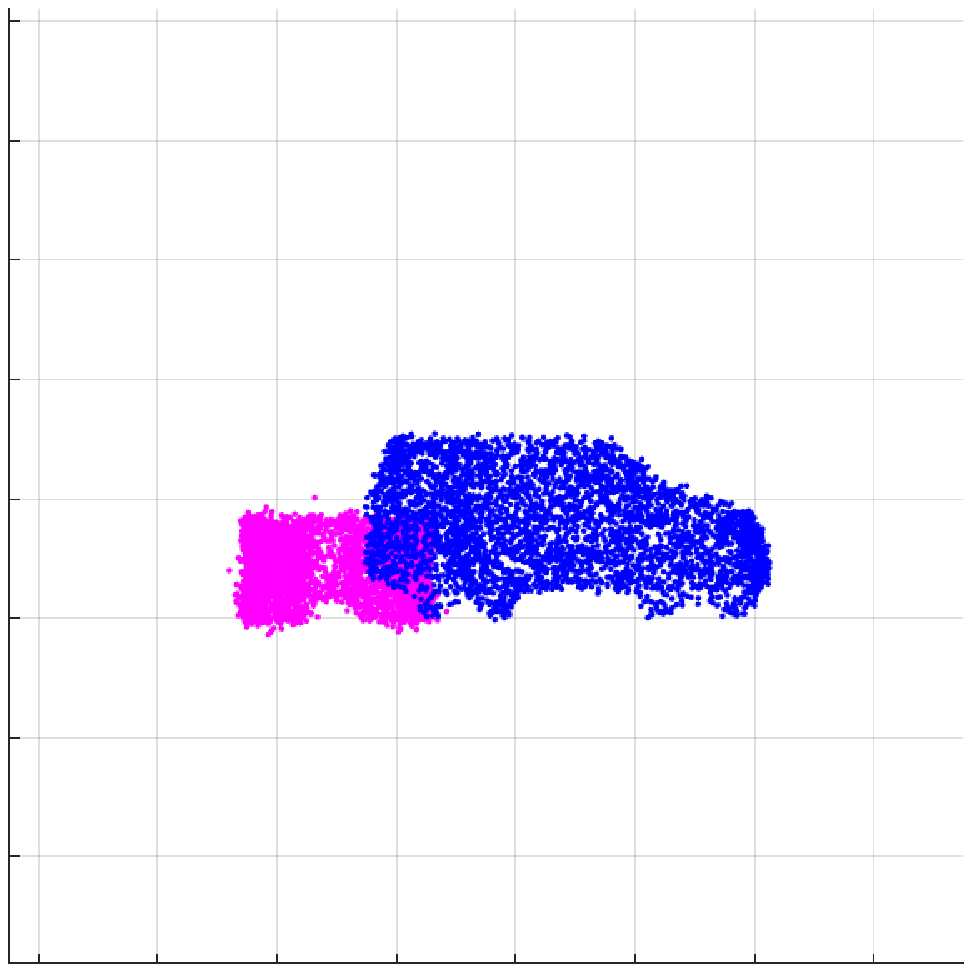}}
		\end{minipage}
		&
		\begin{minipage}[b]{0.4\columnwidth}
			\centering
			\raisebox{-.5\height}{\includegraphics[width=\linewidth]{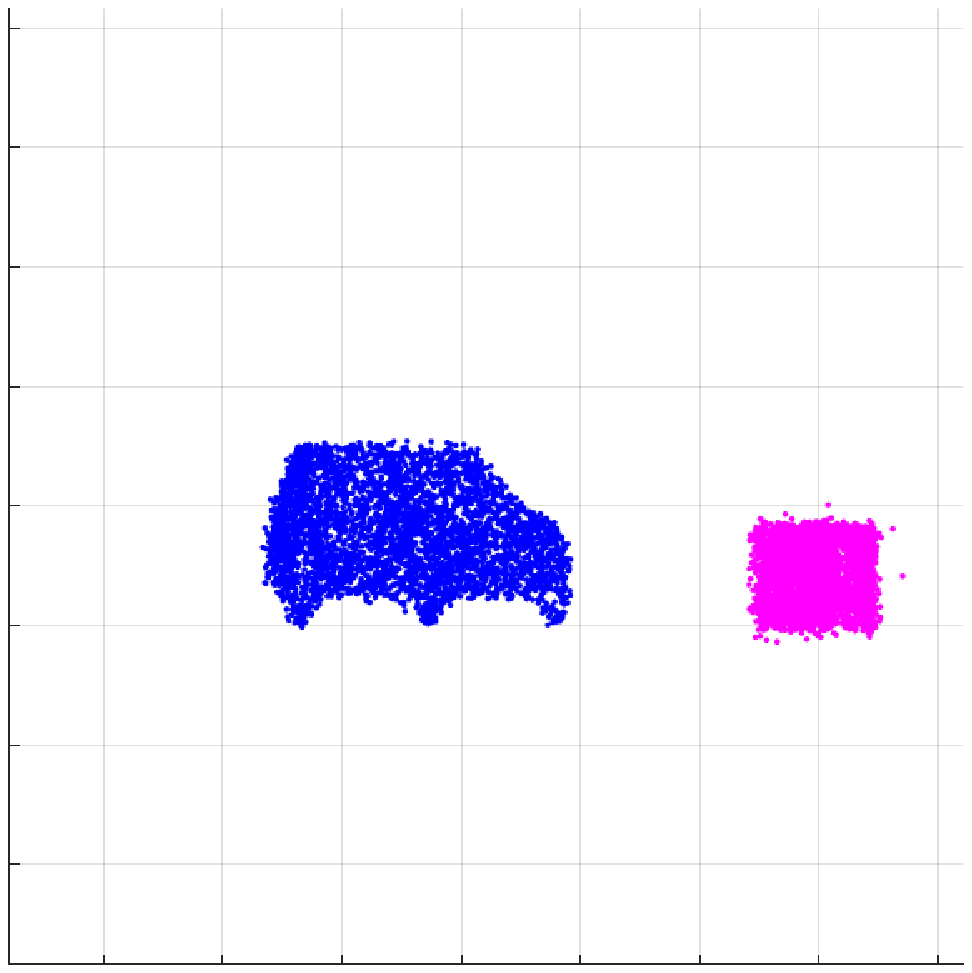}}
		\end{minipage}
		&
		\begin{minipage}[b]{0.4\columnwidth}
			\centering
			\raisebox{-.5\height}{\includegraphics[width=\linewidth]{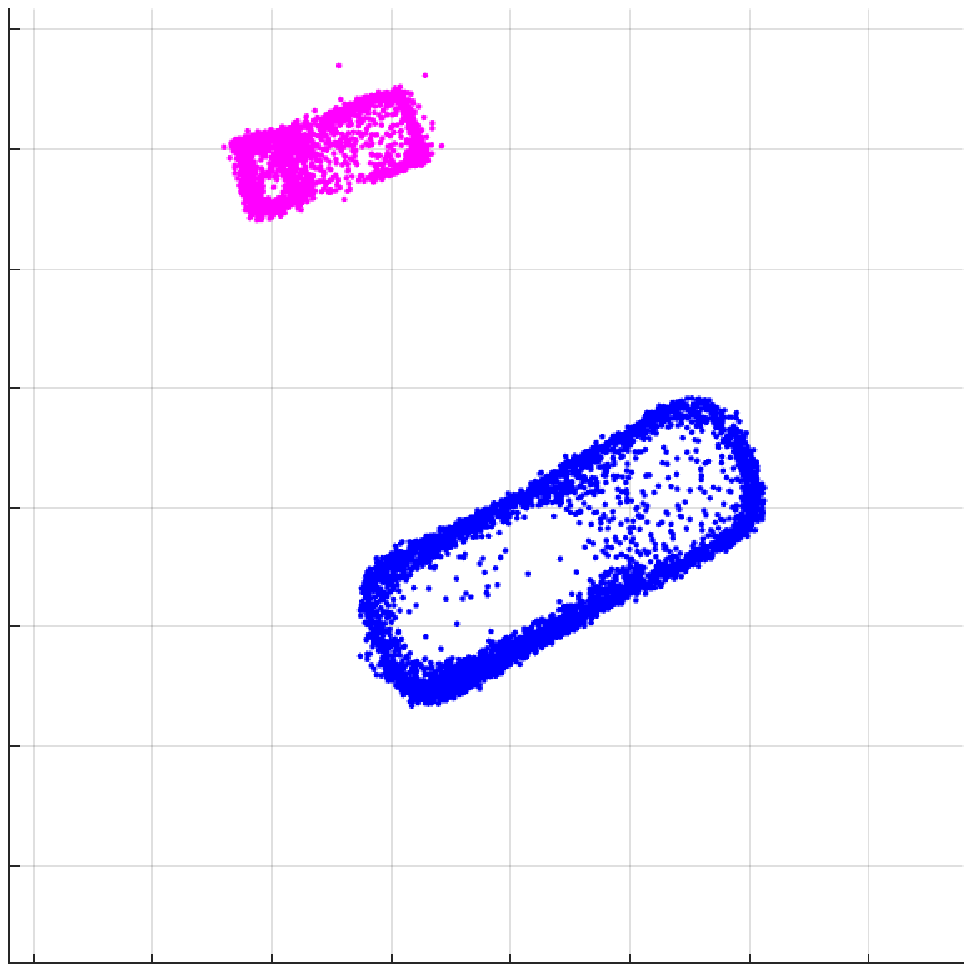}}
		\end{minipage}
		\\
		% row: Ground Truth ---------------------------------------------------------------------------------------
		Ground Truth
		&
		\begin{minipage}[b]{0.4\columnwidth}
			\centering
			\raisebox{-.5\height}{\includegraphics[width=\linewidth]{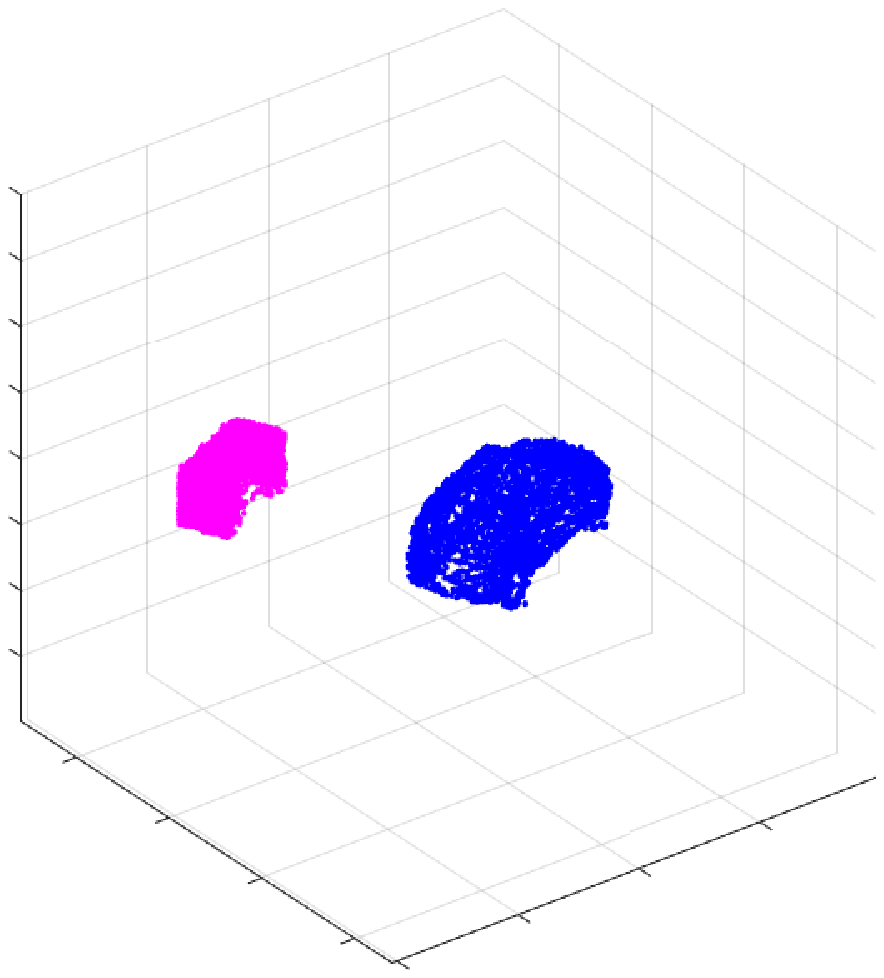}}
		\end{minipage}
		&
		\begin{minipage}[b]{0.4\columnwidth}
			\centering
			\raisebox{-.5\height}{\includegraphics[width=\linewidth]{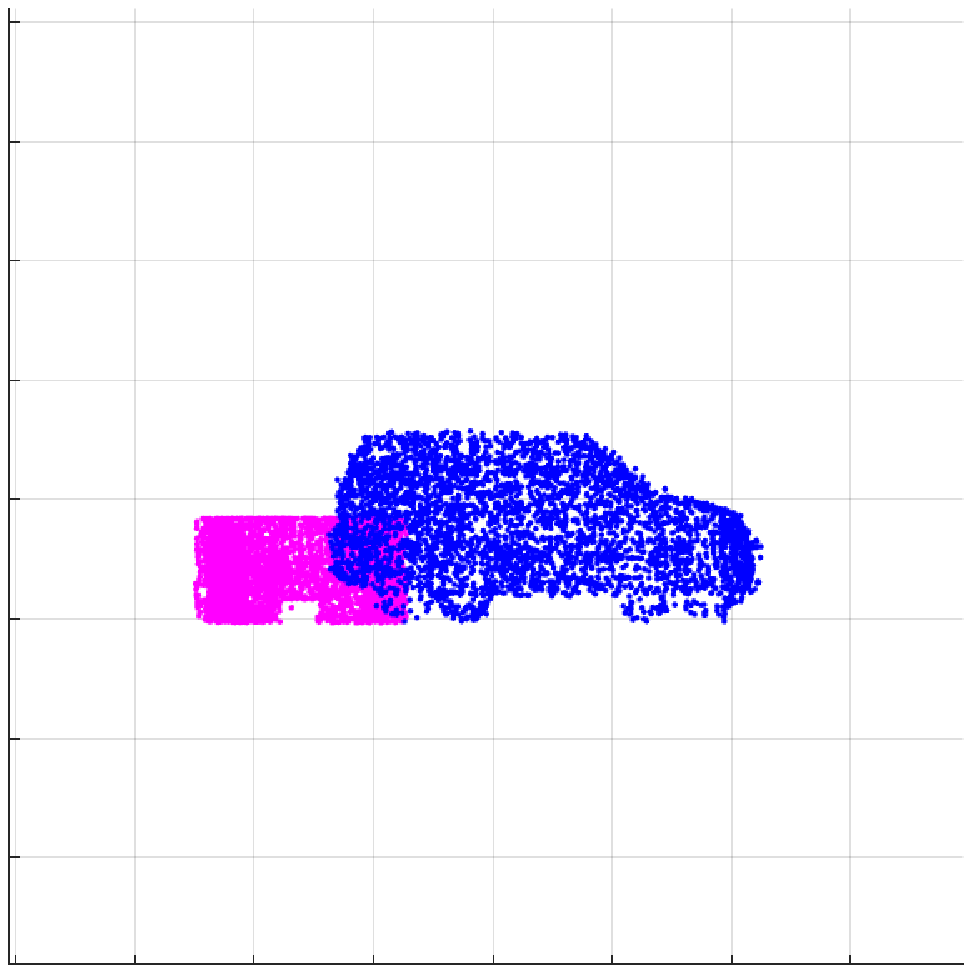}}
		\end{minipage}
		&
		\begin{minipage}[b]{0.4\columnwidth}
			\centering
			\raisebox{-.5\height}{\includegraphics[width=\linewidth]{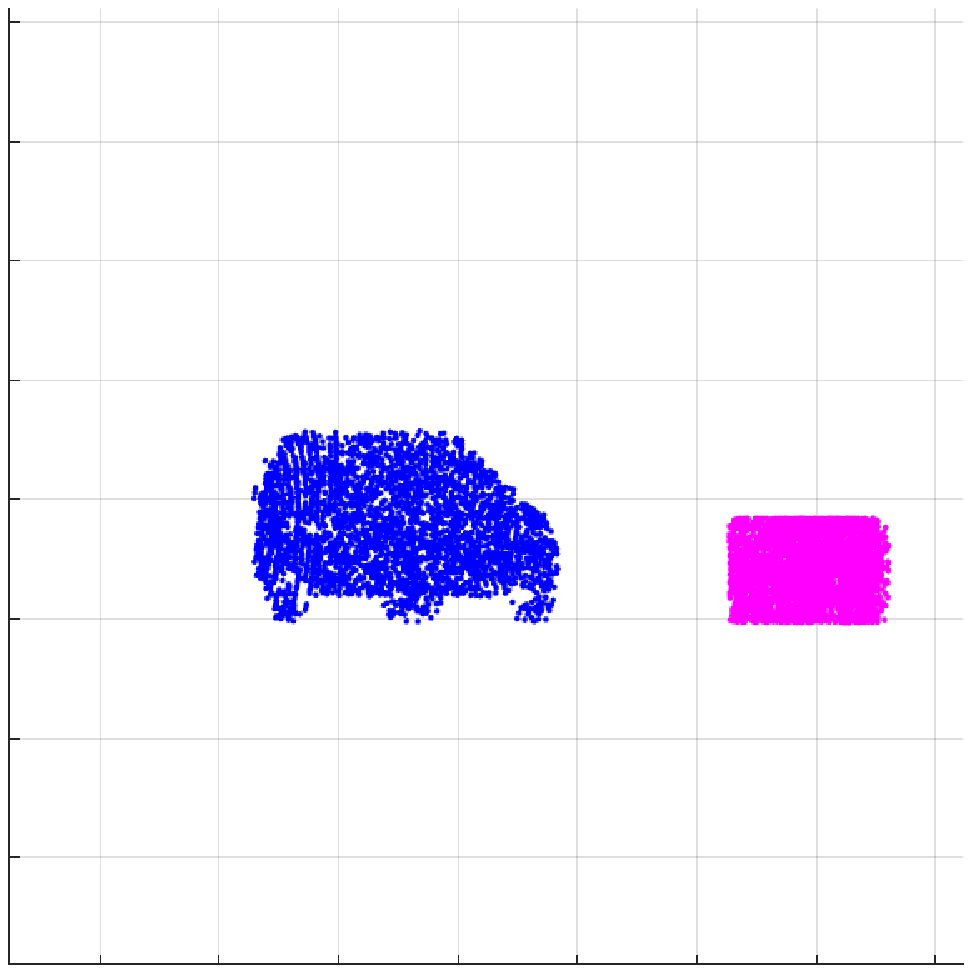}}
		\end{minipage}
		&
		\begin{minipage}[b]{0.4\columnwidth}
			\centering
			\raisebox{-.5\height}{\includegraphics[width=\linewidth]{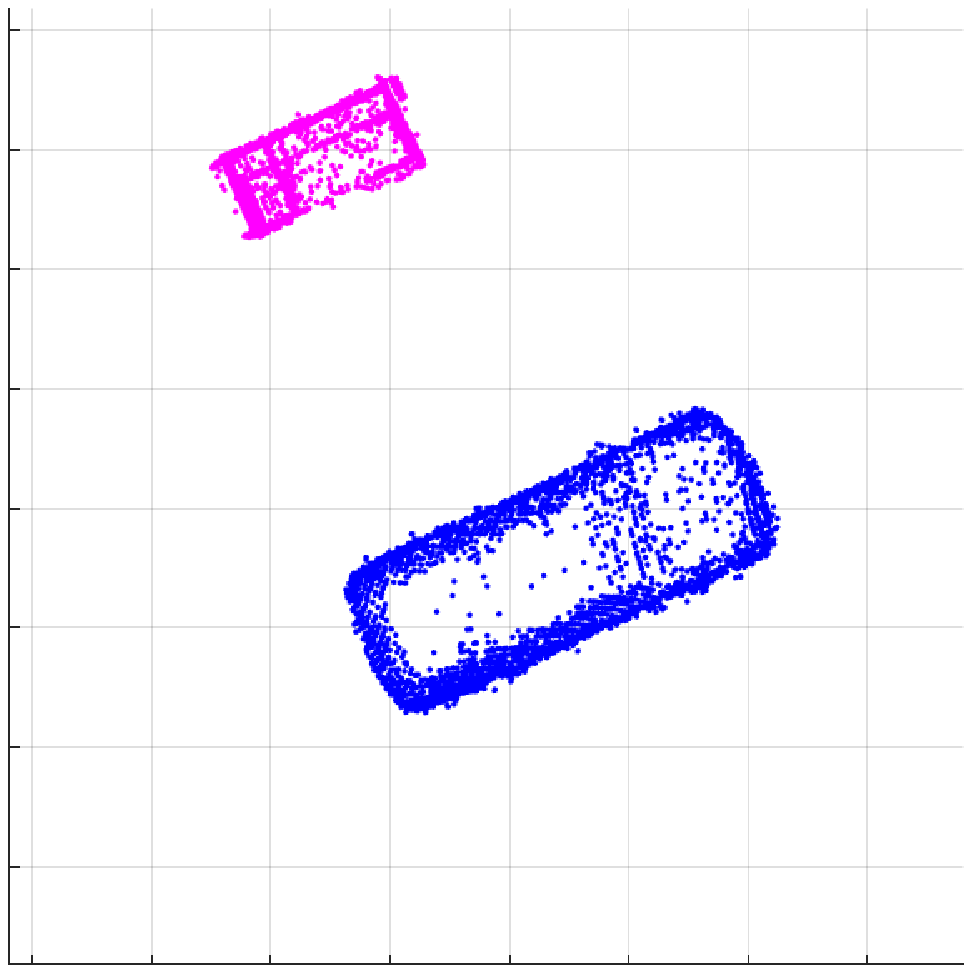}}
		\end{minipage}	
	\end{tabular}
	\captionof{figure}{Comparison of generated point clouds of the scene with 2 objects (a car and a desk) using different methods. }
	\label{pc_plots}
\end{table*}

% ================================================================================================================
\section{CONCLUSIONS AND FUTURE WORK}\label{sec_conclusion}
We have shown via an exploratory study that  
it is feasible to utilize the mmWave radar data collected by a vibrating small UAV's unstable SAR operation to reconstruct 3D shapes of multiple objects in a space. 
We have studied two deep neural network models, 
%for such reconstruction purposes. 
and our experiments results have shown that they achieve promising results 
and they are robust to the vibrating UAV's SAR operation.
%and our experiments have shown that this indeed is a promising approach and 
%the simple models studied in this paper are robust to unstable UAV SAR operation.
%, and the they can produce reasonable good reconstruction results. 
For future work, we will develop novel modules into the architectures of these models
to significantly improve their performance, and we will conduct large scale 
real-world experiments to develop an extensive dataset to test our models.

%
% REFERENCES
%
\newpage

\bibliographystyle{IEEEtran}
\bibliography{IEEEabrv,references}

\end{document}